%% file: main.tex
\input{preamble}
\title{\method: Learning from Visibility Asymmetry for Unsupervised Stereo Matching}
\author{Peng Xu \qquad Zhiyu Xiang\thanks{Corresponding author.}\\
\normalfont College of Information Science and Electronic Engineering\\
\normalfont Zhejiang University, China}

\begin{document}
\maketitle
\lhead{Preprint}
\input{sec/0_abstract}
\begingroup
\setlength{\intextsep}{0pt}
\input{teaser}
\FloatBarrier
\endgroup
\input{sec/1_intro}
\input{sec/2_related_work}
\input{sec/3_method}

\input{sec/4_experiments}
\input{sec/5_conclusion}

\FloatBarrier
\clearpage
\input{ai_use_statement}
\input{reproducibility_statement}
\bibliographystyle{iclr2027_conference}
\bibliography{magicup}

\clearpage
\appendix
\section*{Appendices}
\input{sec/6_supp}
\end{document}

%% file: preamble.tex
\documentclass{article}
\usepackage{iclr2027_conference,times}
\usepackage[hyphens]{url}
\usepackage{graphicx}
\usepackage{booktabs}
\usepackage{amsmath}
\usepackage{amssymb}
\usepackage{multirow}
\usepackage[table]{xcolor}
\usepackage{pifont}
\usepackage{xspace}
\usepackage{hyperref}
\usepackage{placeins}
\usepackage[capitalise]{cleveref}
\usepackage{pgfplots}
\usepackage{tikz}
\pgfplotsset{compat=1.17}
\definecolor{kt15blue}{RGB}{40,74,138}
\definecolor{kt12blue}{RGB}{140,185,225}

\newcommand{\method}{MAGIC\xspace}
\newcommand{\dataset}{MBS20K\xspace}

\DeclareMathOperator{\sg}{sg}

\newcommand{\eg}{\emph{e.g}\onedot}

\makeatletter
\def\onedot{.\@\xspace}
\makeatother

\definecolor{firstcolor}{HTML}{BDE6CD}
\definecolor{secondcolor}{HTML}{E2EEBC}
\definecolor{thirdcolor}{HTML}{FFF8C5}

\iclrfinalcopy

\newcommand{\tablefont}{\fontsize{8}{10}\selectfont}

\newcommand{\panelcaption}[1]{\par\vspace{4pt}{\small #1\par}}

\AddToHook{env/table/begin}{\setlength{\abovecaptionskip}{0pt}}
\AddToHook{env/table*/begin}{\setlength{\abovecaptionskip}{0pt}}

\hypersetup{pdftitle={MAGIC: Learning from Visibility Asymmetry for Unsupervised Stereo Matching},pdfauthor={Peng Xu and Zhiyu Xiang}}

%% file: sec/0_abstract.tex
\begin{abstract}
Learning disparity in occluded regions remains difficult for unsupervised stereo matching.
Photometric supervision lacks valid target-view correspondences in these regions, while the teacher and student in conventional binocular self-training share the same target view and therefore the same occlusions.
Even when supervision is available, the small proportion of occluded pixels limits their contribution to training.
We propose \textbf{\method}, a multi-baseline geometric consistency framework for reliable and effective occlusion supervision.
The teacher and student share a reference image but use different target views, allowing the teacher to observe correspondences that are occluded from the student.
After aligning disparities across baselines, \method uses predictions from teacher-visible regions to supervise student-occluded regions.
An occlusion-aware weighting strategy strengthens supervision on teacher-visible but student-occluded pixels, preventing their training signal from being overwhelmed by non-occluded regions.
We also introduce \textbf{\dataset}, a synthetic multi-baseline stereo dataset spanning diverse scenes, weather, and lighting.
Pre-trained on \dataset, \method generalizes to real-world datasets with consistently fewer occluded-region outliers.
On KITTI, the pre-trained model already outperforms several fine-tuned unsupervised methods.
Fine-tuning this model on standard binocular pairs achieves state-of-the-art unsupervised performance on KITTI 2015 and 2012.
Incorporating synthesized multi-baseline views during fine-tuning further improves performance.
Our code and dataset will be released upon acceptance.
\end{abstract}

%% file: teaser.tex
\begin{figure}[!ht]
    \centering
    \includegraphics[width=\linewidth]{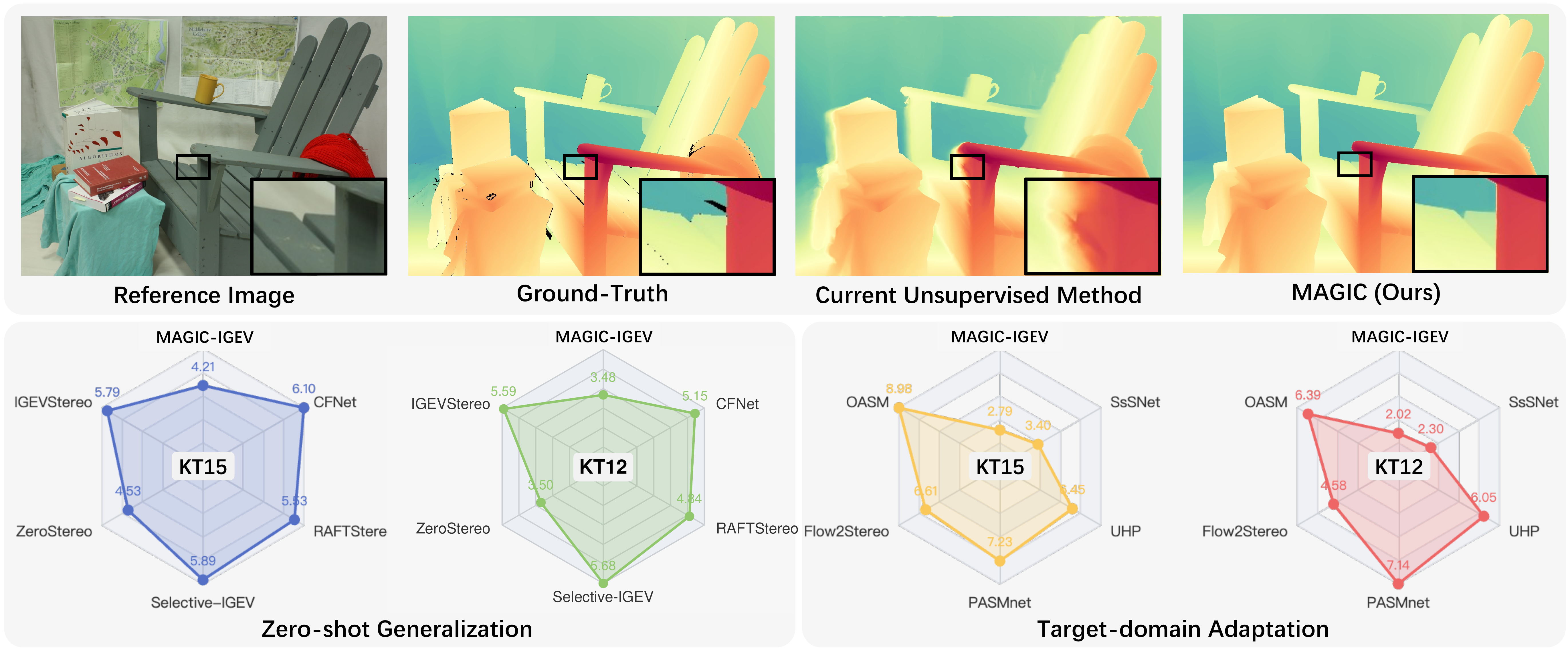}
    \par\vspace{-11pt}
    \caption{
        \textbf{Top:} \method produces accurate disparity estimates in occluded regions (black box; please zoom in for more details).
        \textbf{Bottom:} \method achieves outstanding performance in both zero-shot generalization (left) and target-domain adaptation (right) on KITTI 2015 and 2012~\citep{KITTI2015,KITTI2012} (lower is better).
    }
    \label{fig: teaser}
\end{figure}

%% file: sec/1_intro.tex
\section{Introduction}

Stereo matching recovers dense correspondences between rectified image pairs and serves as a fundamental component in applications such as autonomous driving, robotics, and augmented reality. As ground-truth disparities are costly to acquire, unsupervised methods have attracted increasing attention by learning directly from unlabeled image pairs, most commonly through photometric consistency between the reference and target views~\citep{SsSnet}.

This correspondence-based supervision breaks down in occluded regions\footnote{We treat a reference pixel as \textit{occluded} if its corresponding scene point is hidden behind another surface in the target view or lies outside the target camera's field of view.}. Without a valid correspondence in the target image, disparity must be inferred from contextual cues, making its estimation a \textit{completion} task. Applying the photometric loss in these regions misleads learning (\cref{fig: teaser}). A common remedy detects occlusions and excludes them from the loss~\citep{OASM}. Although this suppresses erroneous gradients, it leaves disparity completion under-constrained.

Pseudo-label self-training~\citep{ZOLE} provides an alternative source of supervision but shares the same visibility limitation under a standard binocular setup. A teacher network predicts pseudo-disparities for the student, yet both networks observe the same target view. Regions occluded to the student therefore remain invisible to the teacher. The teacher's pseudo-labels can be unreliable in occluded regions, and their errors can propagate to the student during training.

A second challenge remains even when reliable occlusion supervision is available. Occluded pixels constitute only a small fraction of an image, so their learning signal can be overwhelmed by the abundant non-occluded regions. Effective occlusion learning therefore requires not only reliable supervision, but also a mechanism that prevents such supervision from being diluted (\cref{fig: occlusion}).

To obtain reliable supervision in occluded regions, we revisit a common assumption in self-training that the teacher and student must observe the same target view. When additional viewpoints with varying baselines are available, a region occluded in one view is frequently visible in another. Moreover, as long as both networks share the same reference image, the underlying scene geometry remains consistent, so disparities predicted under different baselines can be aligned through a simple scaling relation. This visibility asymmetry enables the teacher to rely on valid correspondence cues in its own view and to provide reliable pseudo-labels for regions that are occluded to the student.

Building on this observation, we propose \method, a multi-baseline geometric consistency framework for unsupervised stereo matching. \method retains the teacher-student paradigm~\citep{dino} but assigns distinct target views to the two networks, and rescales the teacher's disparities by the baseline ratio to obtain pseudo-labels that satisfy geometric consistency for the student. To address the supervision imbalance, we further introduce an occlusion-aware weighting strategy that detects occluded regions in the teacher and student views, then amplifies supervision on teacher-visible yet student-occluded pixels and discards the teacher's unreliable pseudo-labels.

To support the training of \method, we construct \dataset, a synthetic multi-baseline stereo dataset rendered in the CARLA simulator~\citep{carla} using a five-camera rig under diverse weather and illumination conditions. Its 20K synchronized frame sets can be combinatorially expanded into millions of training triplets.

Pre-trained on \dataset, \method exhibits strong zero-shot generalization across diverse real-world scenes and weather conditions, reducing occluded-region outliers on KITTI 2015 and 2012 by 27.3\% and 36.2\%, respectively. When submitted to the KITTI benchmarks without any fine-tuning, it already outperforms several fine-tuned unsupervised methods. After fine-tuning with only standard binocular pairs, \method establishes a new unsupervised state-of-the-art, owing to its robust pre-training, and improves further when multi-baseline observations are available in the target domain.
Our contributions are summarized as follows:
\begin{itemize}
\item We propose \method, a teacher--student framework that exploits visibility asymmetry and transfers disparity supervision across different baselines through baseline-ratio rescaling.
\item We identify supervision imbalance as a key factor hindering learning in occluded regions and address it with an occlusion-aware weighting strategy that upweights sparse occluded pixels to learn accurate disparity completion in these regions.
\item We construct \dataset, a synthetic multi-baseline stereo dataset for pre-training with geometric consistency.
\item In the zero-shot setting, \method generalizes strongly; after binocular fine-tuning, it establishes a new unsupervised state-of-the-art on KITTI 2015 and 2012, with further gains from multi-baseline target-domain data.
\end{itemize}

%% file: sec/2_related_work.tex
\section{Related Work}

\noindent\textbf{Deep stereo matching.}
Stereo matching has evolved over the past decades from hand-crafted pipelines~\citep{1} to end-to-end deep learning frameworks~\citep{SceneFlow}.
Modern deep stereo networks generally fall into two architectural families: 3D convolution-based cost volume regularization~\citep{GCNet,GwcNet,CFNet} and ConvGRU-based iterative refinement~\citep{RAFTStereo,CREStereo,dlnr}.
IGEVStereo~\citep{igev} employs a lightweight 3D convolutional module to provide strong geometric initialization for iterative refinement, and Selective-Stereo~\citep{wang2024selective} aggregates multi-frequency representations to improve performance in edge and textureless regions.
More recent approaches~\citep{defomstereo,monster,foundationstereo,wen2026fast} further incorporate monocular depth priors derived from foundation models~\citep{depthanythingv2}.

\noindent\textbf{Photometric-based supervision.}
Photometric consistency provides a natural supervisory signal for training stereo networks on unlabeled data~\citep{SsSnet}.
To alleviate photometric failure in occluded regions, OASM~\citep{OASM} masks photometrically inconsistent pixels, PASMnet~\citep{PASMnet} suppresses unreliable regions via attention-based confidence filtering, and UHP~\citep{UHP} incorporates planar geometric priors.
However, masking occluded areas removes their direct supervision, leaving disparity completion under-constrained.

\noindent\textbf{Pseudo-label-based self-training.}
Self-training supervises students with teacher pseudo-disparities.
ZOLE~\citep{ZOLE} exploits zoomed-in patches for additional image detail, and DualNet~\citep{dualnet} uses feature-metric and augmentation consistency to train a teacher, then freezes it for student supervision.
Other methods update teachers via an exponential moving average (EMA)~\citep{BYOL}. \citet{zhang2024robusttransfer} combine frozen and EMA teachers with ground truth and pseudo-labels to preserve generalization during stereo fine-tuning. TS-PSMNet~\citep{semi-stereo} filters uncertain predictions by confidence, while CST-Stereo~\citep{cst-stereo} extracts reliable pseudo-labels through multi-resolution and iterative consistency.
With the same target viewpoint, however, both networks share the same occlusions, leaving the teacher without additional correspondence cues where the student is occluded.

\noindent\textbf{Multi-view supervision.}
NeRFStereo~\citep{tosi2023nerf} uses a central reference and two target views, taking the per-pixel minimum reconstruction error to alleviate occlusion.
SMV~\citep{multiscopic} aligns center-left and center-right disparities through uncertainty gating, allowing the confident view to correct the other in occluded regions.
Both methods use equal baselines. In self-supervised multi-view stereo (MVS), KD-MVS~\citep{ding2022kdmvs} validates depths across views before probabilistic teacher--student distillation.
For stereo disparity transfer, \method rescales teacher predictions by the baseline ratio, removing the equal-baseline constraint and combinatorially exploiting multi-baseline data.
Moreover, \method applies occlusion-aware weighting to strengthen supervision in occluded regions, whose sparse signal is otherwise overwhelmed by supervision from the abundant non-occluded pixels throughout the training images.

%% file: sec/3_method.tex
\section{Method}

\subsection{Preliminary}

Given a pair of rectified stereo images, stereo matching estimates, for each pixel in the reference image $I_{ref}$, its disparity $d$ to the corresponding pixel in the target image $I_{tgt}$. The depth is recovered by triangulation as $z = B \cdot f / d$, where $B$ is the baseline and $f$ is the focal length.

Unsupervised methods commonly supervise this prediction through photometric consistency~\citep{SsSnet}. The target image is warped by the predicted disparity to reconstruct the reference view $\hat{I}_{ref}$, and a photometric loss combines SSIM and $L_1$ terms,
\begin{equation}
    \mathcal{L}_{p} = \frac{\alpha}{2}(1-\text{SSIM}(I_{ref}, \hat{I}_{ref})) + (1-\alpha) ||I_{ref} - \hat{I}_{ref}||_1.
    \label{eq: photometric}
\end{equation}
To complement photometric supervision, an edge-aware smoothness term $\mathcal{L}_s$ encourages locally smooth disparities while preserving discontinuities at image edges~\citep{SsSnet}.

Another line of unsupervised methods uses a teacher to supervise a student with pseudo-disparities~\citep{ZOLE}. Throughout, superscripts $s$ and $t$ denote the student and teacher, respectively. Given their predicted disparities $d^s$ and $d^t$, the student is supervised by $|d^s - \sg(d^t)|$, where $\sg(\cdot)$ denotes the stop-gradient operator applied to the teacher prediction.

However, neither approach provides reliable supervision in occluded regions. Photometric supervision lacks valid correspondences in these regions. In binocular self-training, the teacher also lacks matching evidence and must solve the more challenging task of disparity completion, potentially producing unreliable pseudo-labels that propagate errors to the student.

\subsection{Multi-Baseline Geometric Consistency}

\method is an unsupervised framework that exploits geometric consistency across stereo observations captured with different reference-to-target camera baselines.

\begin{figure}[t]
    \centering
    \includegraphics[width=\linewidth]{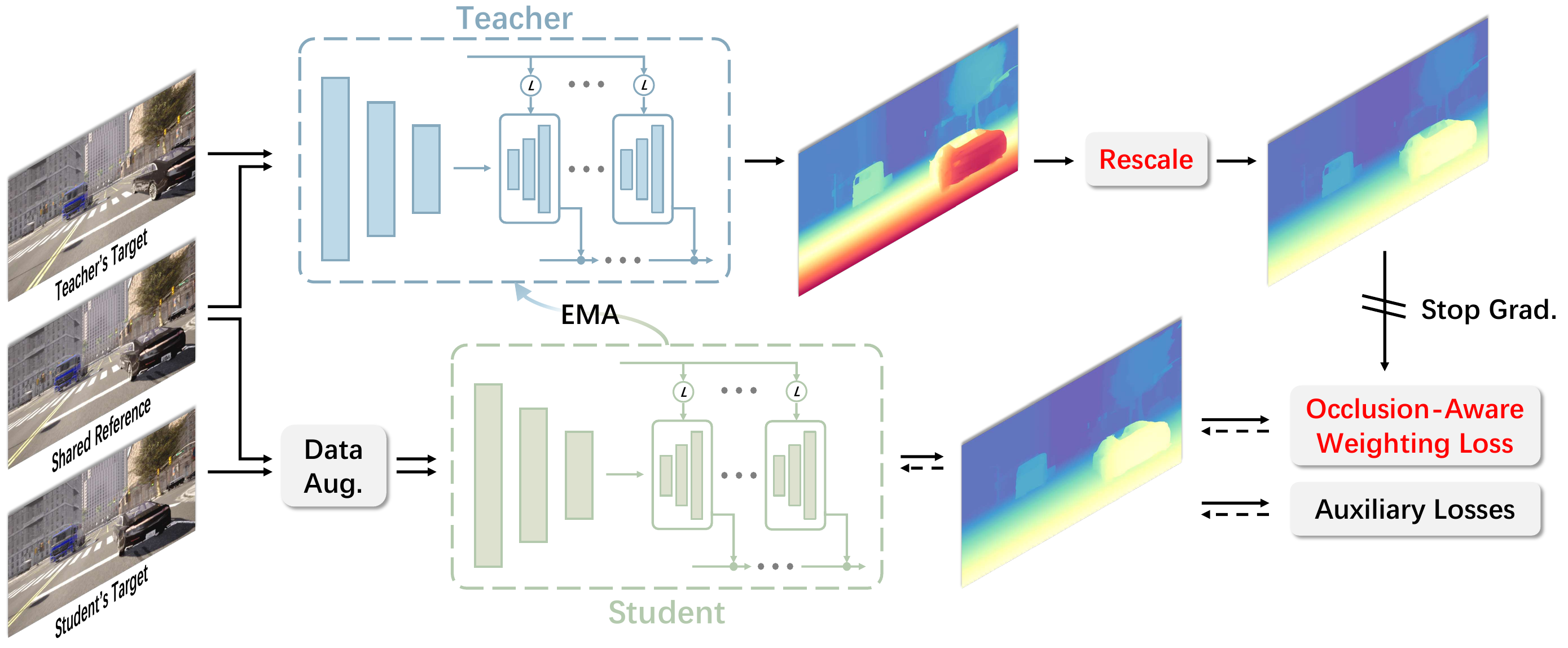}
    \par\vspace{-7pt}
    \caption{
        \textbf{Overview of \method.}
        The illustrated input triplet shares a reference image but uses different target views, letting the teacher observe regions occluded in the student's view.
        Teacher disparities are rescaled by the baseline ratio to match the student's disparity scale. The resulting pseudo-labels are detached from the computational graph and used to supervise the student. Occlusion-aware weighting strengthens supervision on sparse occluded pixels so that their contribution is not overwhelmed by abundant non-occluded pixels.
        Data augmentation is applied only to the student's input, while the teacher is updated as an EMA of the student's weights.}
    \label{fig: framework}
\end{figure}

\textbf{Asymmetric input viewpoints.}
Based on the teacher-student paradigm, \method introduces visibility asymmetry.
As illustrated in~\cref{fig: framework}, the student and teacher networks share the same reference image $I_{ref}$ but receive independently sampled target views $I_{tgt}^s$ and $I_{tgt}^t$.
Different target views yield complementary visibility patterns. Pixels occluded in the student's target view may remain visible in the teacher's view.
Relying on valid correspondence cues available in its own view, the teacher can predict accurate disparities for these student-occluded regions.

Since both networks operate on the same reference image, the underlying 3D scene geometry remains consistent between them.
For a pixel with depth $z$ in the reference image, the predicted disparities $d^s$ and $d^t$ are related by
\begin{equation}
\frac{B^s \cdot f}{d^s} = \frac{B^t \cdot f}{d^t} = z
\quad\Rightarrow\quad
d^s = r \cdot d^t,\;
r = \frac{B^s}{B^t}.
\end{equation}
Rescaling the teacher's disparity by the baseline ratio $r$ maps the teacher's prediction into the student's disparity space, enabling supervision to be transferred across baselines.

\textbf{Robust student and momentum teacher.}
\method also introduces appearance asymmetry through student-only augmentation, including color jitter and random occlusion.
These more challenging student inputs encourage robust features for disparity estimation~\citep{depthanything}.

The teacher is initialized from the student weights and updated using an EMA of the student's weights during training.
The update rule is
$\theta^t \leftarrow m \cdot \theta^t + (1-m) \cdot \theta^s$,
where $\theta$ denotes the model weights, and $m$ is a momentum parameter following a cosine schedule~\citep{BYOL}.
Empirically, a momentum teacher consistently improves performance compared to a fixed teacher.
During inference, only the teacher is used for disparity estimation.

\textbf{Geometric consistency loss.}
The primary supervision signal enforces consistency between the student's prediction and the rescaled teacher disparity:
\begin{equation}
    \mathcal{L}_g = \left\| d^s - r \cdot \sg(d^t) \right\|_1.
\end{equation}
Unlike binocular self-training, this geometric consistency objective reliably supervises student-occluded regions by transferring geometrically aligned predictions from teacher-visible regions.

\begin{figure}[t]
    \centering
    \includegraphics[width=1\linewidth]{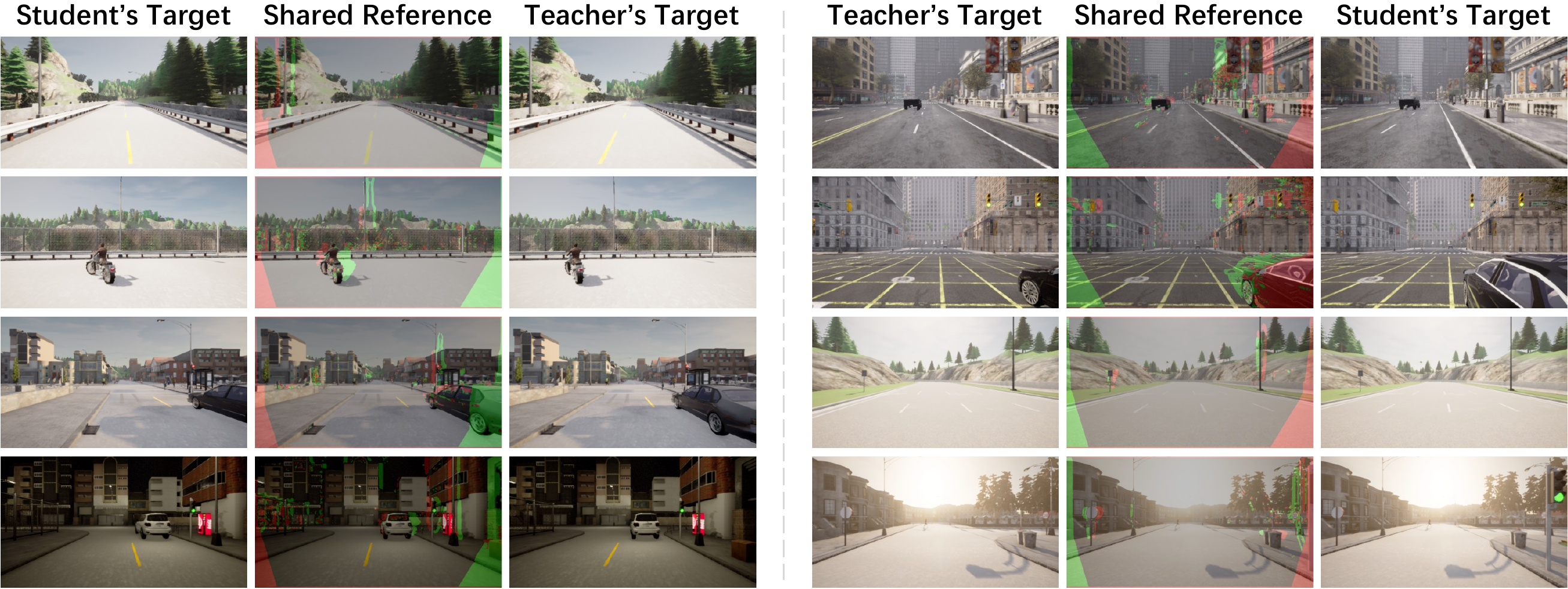}
    \par\vspace{-7pt}
    \caption{
    \textbf{Visualization of the occlusion-aware weight map.}
        We show two triplet arrangements. Colors indicate the visibility regions defined in~\cref{sec3.3}.
        Red areas denote \emph{teacher-occluded} regions, where supervision is discarded due to unreliable teacher predictions.
        Green areas denote \emph{student-occluded yet teacher-visible} regions, where supervision is emphasized.
        The remaining areas denote \emph{co-visible} regions.
        To highlight the visibility differences between the teacher and student views, auto-masking~\citep{monodepth2} is disabled here.
    }
    \label{fig: weighting map}
\end{figure}

\subsection{Occlusion-Aware Weighting}
\label{sec3.3}

For distinct target views, multi-baseline transfer involves three visibility regions in the reference image:
(1) \emph{co-visible} regions, where both the student and teacher observe valid correspondences;
(2) \emph{teacher-occluded} regions, where the teacher lacks reliable correspondence cues; and
(3) \emph{student-occluded yet teacher-visible} regions, where supervision for the student's disparity completion becomes uniquely available through visibility asymmetry.

Region (2) introduces unreliable pseudo-labels, whereas region (3) provides teacher guidance where student correspondences fail.
Appendix~\ref{sec:weight_maps} provides an analysis of the effectiveness of teacher pseudo-labels in region (3).
However, obtaining this supervision does not ensure that it contributes effectively to learning.
Neural networks tend to prioritize simple and frequent patterns~\citep{2}, so uniform weighting allows abundant co-visible pixels to dominate optimization, limiting disparity completion learning in the sparse region (3).

\suppressfloats[t]
\begin{table}[t]
    
    \centering
    \caption{\textbf{Zero-shot generalization across diverse scenes.} We report Out-x in occluded (OCC), non-occluded (NOC), and all (ALL) regions on four real-world datasets. $\dagger$ and $\ddagger$ denote methods that incorporate real data or foundation priors, respectively.}
    \label{tab: scene generalization}
    \par\vspace{2pt}
    \tablefont
    \setlength{\tabcolsep}{3pt}
    \begin{tabular}{lcccccccccccc}
         \toprule
         \multirow{2}{*}{Method} & \multicolumn{3}{c}{KITTI 2015} & \multicolumn{3}{c}{KITTI 2012} & \multicolumn{3}{c}{Middlebury} & \multicolumn{3}{c}{ETH3D} \\
         \cmidrule(lr){2-4} \cmidrule(lr){5-7} \cmidrule(lr){8-10} \cmidrule(lr){11-13}
         & OCC & NOC & ALL & OCC & NOC & ALL & OCC & NOC & ALL & OCC & NOC & ALL \\
         \midrule
        PSMNet~(\citeyear{PSMNet}) & 47.64 & 28.13 & 28.42 & 63.20 & 26.50 & 27.32 & 62.30 & 30.18 & 34.51 & 28.56 & 14.74 & 15.39 \\
        GwcNet~(\citeyear{GwcNet}) & 29.07 & 12.17 & 12.53 & 45.65 & 11.91 & 12.67 & 47.13 & 20.41 & 24.11 & 21.37 & 10.49 & 11.09 \\
        CFNet~(\citeyear{CFNet}) & 16.42 & 5.87 & 6.10 & 30.25 & 4.58 & 5.15 & 44.55 & 16.33 & 20.22 & 11.89 & 5.57 & 5.87 \\
        RAFTStereo~(\citeyear{RAFTStereo}) & 12.70 & 5.34 & 5.53 & 28.35 & 4.29 & 4.84 & 28.00 & 9.06 & 11.96 & 6.02 & 2.85 & 3.04 \\
        Selective-IGEV~(\citeyear{wang2024selective}) & 13.82 & 5.70 & 5.89 & 31.85 & 5.06 & 5.68 & 22.59 & 6.73 & 9.17 & 9.81 & 4.07 & 4.43 \\
        NeRFStereo$^{\dagger}$~(\citeyear{tosi2023nerf}) & 14.62 & 5.23 & 5.43 & 26.97 & 3.51 & 4.04 & 31.10 & 6.79 & 10.38 & 8.35 & 2.78 & 3.07 \\
        ZeroStereo-IGEV$^{\dagger}$~(\citeyear{zerostereo}) & 10.53 & 4.36 & 4.53 & 19.41 & 3.16 & 3.50 & 21.27 & 4.90 & 7.40 & 6.25 & 1.93 & 2.19 \\
        DEFOMStereo$^{\ddagger}$~(\citeyear{defomstereo}) & 12.57 & 4.79 & 4.99 & 21.95 & 3.83 & 4.21 & 20.64 & 4.39 & 6.91 & 5.14 & 2.08 & 2.24 \\
        Fast-FoundationStereo$^{\ddagger}$~(\citeyear{wen2026fast}) & 10.18 & 3.39 & 3.57 & 22.10 & 2.78 & 3.19 & 20.41 & 2.33 & 5.05 & 11.26 & 1.21 & 1.72 \\
        IGEVStereo~(\citeyear{igev}) & 14.26 & 5.60 & 5.79 & 33.66 & 4.92 & 5.59 & 24.28 & 7.25 & 9.91 & 9.76 & 4.06 & 4.39 \\
        MonSter$^{\ddagger}$~(\citeyear{monster}) & 9.64 & 3.30 & 3.45 & 18.82 & 3.03 & 3.37 & 18.42 & 5.86 & 7.70 & 3.53 & 1.32 & 1.45 \\
        FoundationStereo$^{\ddagger}$~(\citeyear{foundationstereo}) & \underline{8.10} & 3.07 & 3.20 & \underline{15.28} & 2.63 & 2.91 & \underline{8.53} & \underline{1.12} & \underline{2.26} & \underline{2.65} & \underline{0.52} & \underline{0.61} \\
        \midrule
        \rowcolor{gray!15} \textbf{\method-IGEV} & 10.37 & 4.05 & 4.21 & 21.49 & 3.08 & 3.48 & 23.46 & 6.34 & 8.94 & 7.98 & 2.51 & 2.76 \\
        \rowcolor{gray!15} \textbf{\method-MonSter$^{\ddagger}$} & 8.17 & \textbf{2.53} & \textbf{2.68} & 17.26 & \textbf{2.16} & \textbf{2.48} & 17.78 & 5.59 & 7.13 & 3.47 & 1.19 & 1.35 \\
        \rowcolor{gray!15} \textbf{\method-FoundationStereo$^{\ddagger}$} & \textbf{7.89} & \underline{2.61} & \underline{2.75} & \textbf{13.57} & \underline{2.26} & \underline{2.50} & \textbf{8.43} & \textbf{1.11} & \textbf{2.24} & \textbf{2.58} & \textbf{0.51} & \textbf{0.60} \\
        \bottomrule
    \end{tabular}
\end{table}

\begin{figure}[t]
    \centering
    \includegraphics[width=\linewidth]{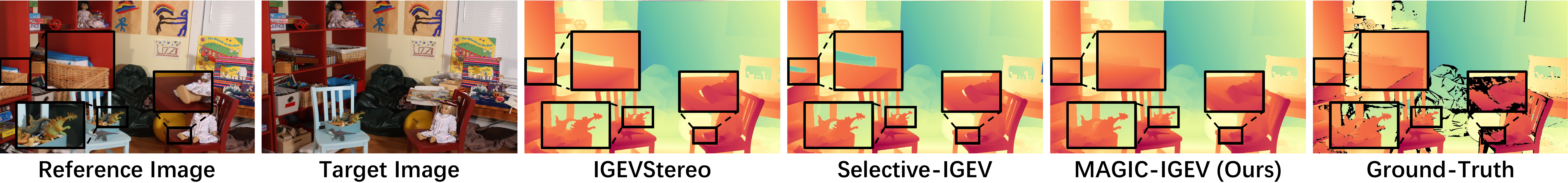}
    \par\vspace{-7pt}
    \caption{\textbf{Qualitative comparison in occluded areas.}
    Despite being trained with full supervision, the competing methods still produce erroneous disparities in occluded regions.
    In contrast, by providing reliable and up-weighted pseudo-supervision, \method improves disparity completion.}
    \label{fig: occlusion}
\end{figure}

To address this supervision imbalance and filter out erroneous guidance, we introduce an occlusion-aware weight map $\mathcal{A}$, defined over the computed teacher and student visibility masks $\mathcal{V}^t,\mathcal{V}^s\in\{0,1\}$, where $\mathcal{V}=1$ indicates a visible pixel.
Each pixel is then assigned a weight as
\begin{equation}
\mathcal{A} = \mathcal{V}^t\big(\mathcal{V}^s + \omega\,(1-\mathcal{V}^s)\big),
\label{eq: mask}
\end{equation}
where unreliable supervision signals from region (2) are explicitly discarded, and a hyperparameter $\omega > 1$ increases the penalty assigned to region (3).
This weighting encourages the student to learn disparity completion from reliable teacher predictions in region (3).

The visibility masks $\mathcal{V}^t$ and $\mathcal{V}^s$ are computed using two complementary criteria.
Since occluded pixels typically incur large photometric errors, we threshold the teacher and student photometric losses, $\mathcal{L}_p^t$ and $\mathcal{L}_p^s$, at $\tau$ to identify likely occlusions in their respective target views.
In addition, we adopt the auto-masking strategy~\citep{monodepth2} to filter out unreliable stationary pixels (in stereo matching, these correspond to infinite-depth regions such as the sky).
A pixel is retained only when it passes both criteria.
Note that $\mathcal{L}_p^t$ is used only for mask computation and does not update the teacher.
\Cref{fig: weighting map} visualizes the occlusion-aware weight maps for the two triplet arrangements, and Appendix~\ref{sec:weight_maps} provides an evaluation of the reliability of the computed visibility masks.

\subsection{Final Loss}
\label{sec3.4}

Optimizing only the baseline-scaled geometric consistency loss $\mathcal{L}_g$ may admit degenerate solutions (\eg, predicting zero disparity, which trivially satisfies the loss without capturing scene structure).
To stabilize optimization and encourage meaningful correspondence learning, we additionally apply the photometric loss $\mathcal{L}_p^s$ and the smoothness loss $\mathcal{L}_s^s$ to the student branch, promoting appearance-based correspondence matching and continuity of local disparity predictions.

The final loss is formulated as
\begin{equation}
    \mathcal{L} = \mathcal{L}_{g}\odot{\mathcal{A}} + \lambda_p\mathcal{L}_{p}^s\odot\mathcal{V}^s + \lambda_s\mathcal{L}_{s}^s ,
    \label{eq: total loss}
\end{equation}
where $\odot$ denotes element-wise multiplication.
The occlusion-aware weight map $\mathcal{A}$ balances supervision across different visibility regions, while $\mathcal{V}^s$ masks unreliable photometric supervision in the student-occluded regions.
Following the loss configuration of~\citep{monodepth2}, we set the weighting parameters to $\alpha=0.85$ in~\cref{eq: photometric} and $\lambda_s = 0.001\lambda_p$ in~\cref{eq: total loss}.

These objectives are complementary: photometric loss establishes basic matching, while teacher supervision progressively improves completion where the student's photometric assumption fails. EMA gradually transfers this improved completion ability back to the teacher. Appendix~\ref{sec:kitti_supp} presents experimental results showing that our framework can be trained from random initialization.

\begin{table}[t]
\begin{minipage}[t]{0.54\linewidth}
\vspace{0pt}

    \centering
    \caption{\textbf{Zero-shot generalization across weather conditions} on DrivingStereo. $\dagger$ and $\ddagger$ denote methods that use real training data or incorporate foundation-model priors, respectively.}
    \label{tab: weather generalization}
    \par\vspace{2pt}
    \fontsize{8}{9}\selectfont
    \setlength{\tabcolsep}{2pt}
    \resizebox{\linewidth}{!}{%
    \begin{tabular}{@{}lrrrr@{}}
         \toprule
         Method & Sunny & Cloudy & Rainy & Foggy \\
         \midrule
         RAFTStereo~(\citeyear{RAFTStereo}) & 4.23 & 4.19 & 12.77 & 3.09 \\
         Selective-IGEV~(\citeyear{wang2024selective}) & 5.05 & 5.24 & 13.51 & 4.10 \\
         NeRFStereo$^{\dagger}$~(\citeyear{tosi2023nerf}) & 2.88 & 2.91 & 10.20 & 3.93 \\
         ZeroStereo-IGEV$^{\dagger}$~(\citeyear{zerostereo}) & 2.59 & 2.36 & 13.07 & \underline{1.89} \\
         DEFOMStereo$^{\ddagger}$~(\citeyear{defomstereo}) & 3.61 & 3.75 & 13.53 & 2.88 \\
         Fast-FoundationStereo$^{\ddagger}$~(\citeyear{wen2026fast}) & 7.08 & 6.96 & 32.65 & 9.83 \\
         IGEVStereo~(\citeyear{igev}) & 4.59 & 5.15 & 15.47 & 4.49 \\
         MonSter$^{\ddagger}$~(\citeyear{monster}) & 3.48 & 3.17 & 5.27 & 5.03 \\
         FoundationStereo$^{\ddagger}$~(\citeyear{foundationstereo}) & 3.97 & 2.73 & 28.50 & 3.11 \\
         \midrule
         \rowcolor{gray!15} \textbf{\method-IGEV} & \underline{2.15} & \underline{1.87} & 5.07 & \textbf{1.64} \\
         \rowcolor{gray!15} \textbf{\method-MonSter$^{\ddagger}$} & \textbf{1.91} & \textbf{1.78} & \textbf{2.35} & 2.11 \\
         \rowcolor{gray!15} \textbf{\method-FoundationStereo$^{\ddagger}$} & 2.20 & 2.27 & \underline{4.18} & 2.18 \\
         \bottomrule
    \end{tabular}}
\end{minipage}\hfill
\begin{minipage}[t]{0.44\linewidth}
    \vspace{0pt}
    \centering
    
    \caption{
        \textbf{Results on KITTI benchmarks}.
        The unmarked \method-IGEV denotes no fine-tuning, $^{*}$ denotes binocular fine-tuning, and $^{\S}$ denotes multi-baseline fine-tuning.
    }
    \label{tab: kitti benchmark}
    \par\vspace{2pt}
    \tablefont
    \setlength{\tabcolsep}{2pt}
    \begin{tabular}{l *{4}{>{\centering\arraybackslash}p{2.4em}}}
        \toprule
        \multirow{2}{*}{Method} & \multicolumn{2}{c}{\shortstack{KITTI 2015\\(D1)}} & \multicolumn{2}{c}{\shortstack{KITTI 2012\\(Out-3)}} \\
        \cmidrule(lr){2-3} \cmidrule(lr){4-5}
        & NOC & ALL & NOC & ALL \\
        \midrule
        SsSnet~(\citeyear{SsSnet})  & 3.06 & 3.40 & 2.30 & 3.00 \\
        OASM~(\citeyear{OASM}) & 7.39 & 8.98 & 6.39 & 8.60\\
        Flow2Stereo~(\citeyear{flow2stereo}) & 6.29 & 6.61 & 4.58 & 5.11 \\
        PASMnet~(\citeyear{PASMnet}) & 6.69 & 7.23 & 7.14 & 8.57\\
        UHP~(\citeyear{UHP}) & 5.93 & 6.45 & 6.05 & 7.09 \\
        DualNet~(\citeyear{dualnet})  & 2.67 & 2.92 & 2.06 & 2.59 \\
        \midrule
        \rowcolor{gray!15} \textbf{\method-IGEV} & 4.27 & 4.60 & 2.72 & 3.27 \\
        \rowcolor{gray!15} \textbf{\method-IGEV}$^{*}$ & \underline{2.59} & \underline{2.79} & \underline{2.02} & \underline{2.36} \\
        \rowcolor{gray!15} \textbf{\method-IGEV$^{\S}$} & \textbf{2.45} & \textbf{2.65} & \textbf{1.92} & \textbf{2.31} \\
        \bottomrule
    \end{tabular}
\end{minipage}
\end{table}

\subsection{Training Data}

For large-scale unsupervised training, we construct a synthetic multi-baseline stereo dataset, \dataset, using the CARLA simulator~\citep{carla}.

\textbf{Dataset overview.}
At each timestamp, a rig of five horizontally aligned RGB cameras captures synchronized images, with a uniform baseline of 0.5\,m between adjacent cameras.
Each camera has a focal length of 480\,px and renders images at a resolution of $540 \times 960$\,px.
In total, 20K multi-baseline image sets are collected across diverse scenes under varying weather and illumination conditions. Appendix~\ref{sec:dataset} provides further dataset details.

\textbf{Sampling strategy.}
During training, one image is randomly selected from the five views as the reference image, while two target views are independently sampled with replacement from the remaining four images.
Unlike conventional binocular stereo setups, target views are not restricted to the right side of the reference image.
If a selected target view lies on the left, the reference and that target are horizontally flipped before being fed into the network, and the predicted disparity is flipped back afterward.
Through this sampling strategy, the 20K multi-baseline image sets can be expanded into 1.6M distinct triplet training samples.

%% file: sec/4_experiments.tex
\section{Experiments}

\subsection{Experimental Setup}

\textbf{Evaluation protocols.}
(1) We pre-train \method on \dataset without disparity labels and evaluate its zero-shot generalization on five public real-world datasets: KITTI 2012~\citep{KITTI2012}, KITTI 2015~\citep{KITTI2015}, Middlebury (H)~\citep{Middlebury}, ETH3D~\citep{ETH3D}, and DrivingStereo~\citep{drivingstereo}.
KITTI, Middlebury, and ETH3D provide masks for quantitative occluded-region evaluation.
(2) For a fair comparison with existing unsupervised methods, we further fine-tune the pre-trained model on KITTI 2012 and 2015 using standard binocular image pairs.
Predictions on the test sets are submitted to the online benchmarks.
(3) To evaluate multi-baseline target-domain adaptation, we generate additional viewpoints for the KITTI training data to construct a multi-baseline fine-tuning setting.

\textbf{Evaluation metrics.}
We employ Out-x, which measures the percentage of pixels whose absolute disparity error exceeds $x$\,px.
Following standard practice, we report Out-3 on KITTI and DrivingStereo, Out-2 on Middlebury, and Out-1 on ETH3D.
Additionally, we report the D1 error metric for evaluation on the KITTI 2015 benchmark~\citep{KITTI2015}.

\textbf{Implementation details.}
We conduct experiments with IGEVStereo~\citep{igev}, MonSter~\citep{monster}, and FoundationStereo~\citep{foundationstereo}.
Following~\citep{dualnet}, all networks are initialized with official weights trained on SceneFlow~\citep{SceneFlow}.
For pre-training, models are trained on \dataset for 200K iterations with a batch size of 16.
We use Adam~\citep{adam} with a one-cycle learning rate schedule~\citep{onecyclelr} and a maximum learning rate of $2\times10^{-4}$.
For fine-tuning, models are trained for 50K iterations with the maximum learning rate reduced to $5\times10^{-5}$.
We set $\lambda_p=10$, $\tau=0.1$, and $\omega=2$.
Appendix~\ref{sec:hyperparameter_search} documents the grid search used to select these hyperparameters.

\newcommand{\ctoprule}[1]{%
  \noalign{\vskip\aboverulesep}%
  \cmidrule[\heavyrulewidth]{#1}%
}

\newcommand{\cbottomrule}[1]{%
  \cmidrule[\heavyrulewidth]{#1}%
}

\subsection{Zero-Shot Generalization}
\label{sec:zero-shot-generalization}

To assess the generalization of models pre-trained on synthetic data to unseen real-world environments, we evaluate zero-shot performance across different scenes and weather conditions.

\textbf{Cross-scene evaluation.}
As shown in \cref{tab: scene generalization}, \method-IGEV exhibits robust zero-shot generalization, consistently surpassing its baseline~\citep{igev} across indoor and outdoor scenes.
Applying our framework to the stronger foundation-prior backbones MonSter~\citep{monster} and FoundationStereo~\citep{foundationstereo} further pushes performance to the state-of-the-art, with \method-FoundationStereo ranking first on most metrics.
These results demonstrate the effectiveness of our framework across different stereo backbones.

\textbf{Comparison in occluded regions.}
\looseness=-1
As reported in \cref{tab: scene generalization}, \method-IGEV achieves relative outlier reductions of 27.28\%, 36.16\%, 3.38\%, and 18.24\% on KITTI 2015, KITTI 2012, Middlebury, and ETH3D, respectively.
Qualitative comparisons in \cref{fig: occlusion} further show that existing supervised methods~\mbox{\citep{igev,wang2024selective}} often produce erroneous disparities in occluded regions, whereas our method recovers more coherent structures.
Together, these quantitative and qualitative results show that our framework improves disparity estimation in occluded regions. We additionally analyze the contribution of occlusion-aware weighting in~\cref{sec: ablation}.

\textbf{Comparison with multi-view methods.}
NeRFStereo \citep{tosi2023nerf} and SMV~\citep{multiscopic} also exploit multi-view observations for unsupervised training.
As shown in \cref{tab: scene generalization}, \method-IGEV consistently outperforms NeRFStereo across all datasets, with particularly pronounced improvements in occluded regions.
Since SMV provides no public weights, we follow its evaluation protocol and report zero-shot D1-all on the KITTI 2015 training set, where \method-FoundationStereo achieves 2.57\%, compared with 3.82\% reported by SMV.
\method combines supervision across different baselines with occlusion-aware weighting to improve disparity completion learning in occluded regions.

\textbf{Cross-weather evaluation.}
\cref{tab: weather generalization} reports performance across diverse weather conditions.
Rainy scenes are especially challenging, with FoundationStereo's outlier rate rising to 28.50\%. In contrast, the diversity of our training data keeps \method-FoundationStereo at 4.18\%, a relative reduction of over 85\%.
Furthermore, although fog is absent from our training data, \method improves every backbone by at least 29\%, indicating that our supervision generalizes beyond the training distribution.
Qualitative results in~\cref{fig: rainy_foggy} further illustrate that \method yields stable predictions in challenging regions, including road surfaces covered with water.

\subsection{Target-Domain Adaptation}

\begin{figure}[t]
\begin{minipage}[t]{0.49\linewidth}
\vspace{0pt}

    \centering
    \includegraphics[width=\linewidth]{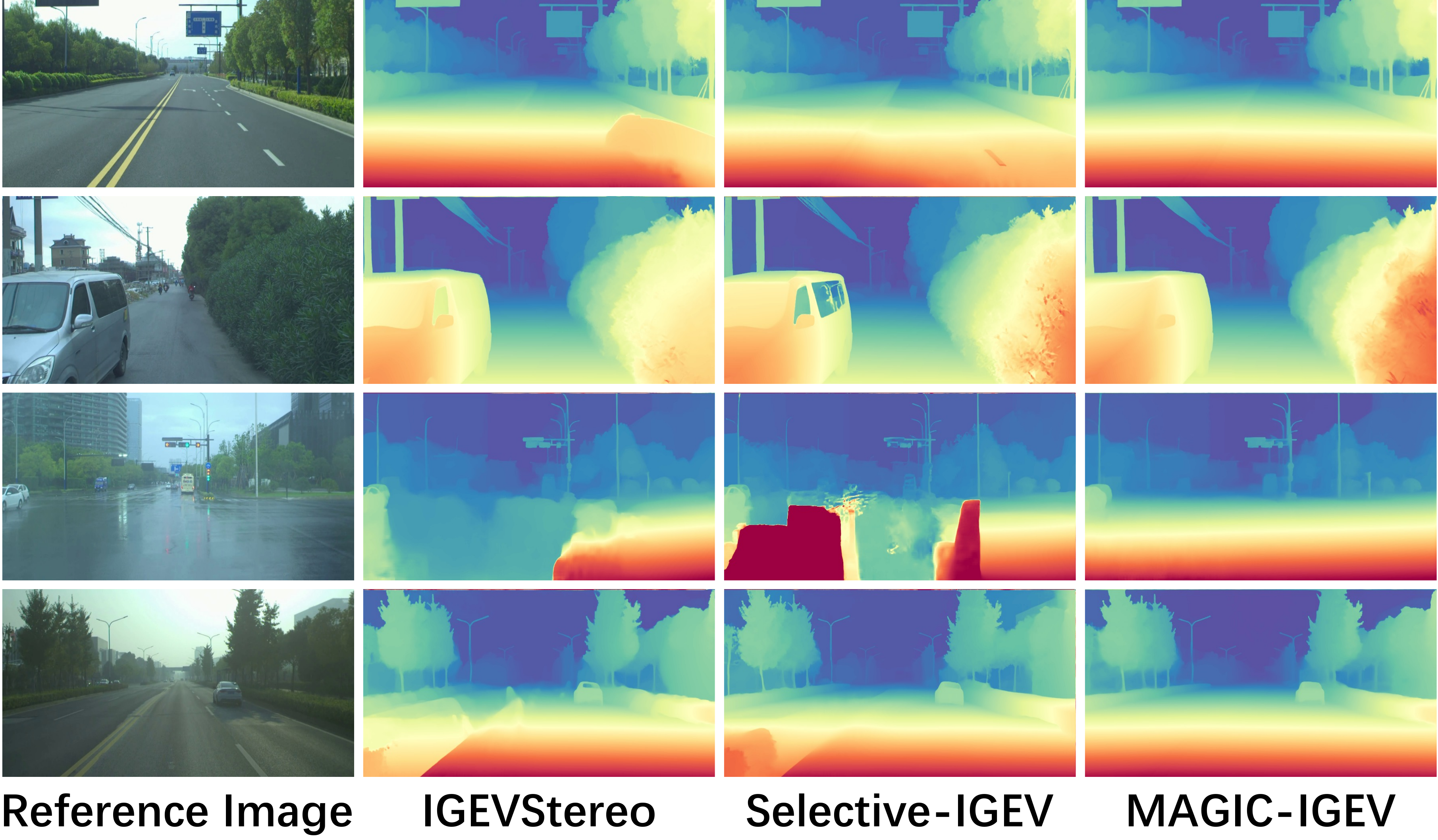}
\end{minipage}\hfill
\begin{minipage}[t]{0.49\linewidth}
    \vspace{0pt}
    \centering
\begin{tikzpicture}
    \begin{axis}[
        width=\dimexpr\linewidth-29pt\relax,
        xbar,
        bar width=7pt,
        y=16.5pt,        %
        enlarge y limits={abs=7.5pt}, %
        xmin=0,
        xmax=11,
        axis y line*=left,
        axis x line=none,
        ytick=data,
        symbolic y coords={
            MAGIC-IGEV,
            AdaStereo,
            TS-PSMNet,
            CST-Stereo,
            ZOLE,
            AoHNet,
            MADNet
        },
        yticklabels={
            {MADNet~(\citeyear{MADNet})},
            {AoHNet~(\citeyear{aohnet})},
            {ZOLE~(\citeyear{ZOLE})},
            {CST-Stereo~(\citeyear{cst-stereo})},
            {AdaStereo~(\citeyear{AdaStereo})},
            {TS-PSMNet~(\citeyear{semi-stereo})},
            {\bfseries MAGIC-IGEV}
        },
        tick style={draw=none},
        every axis plot/.append style={draw=none},
        tick label style={font=\sffamily\fontsize{7}{8.4}\selectfont},
        legend style={
            at={(axis cs:10.9,MAGIC-IGEV)},
            anchor=south east,
            yshift=-7.5pt,
            legend columns=1,
            font=\sffamily\fontsize{7}{8.4}\selectfont,
            draw=none,
            fill=none
        },
        nodes near coords={\pgfplotspointmeta},
        point meta=explicit symbolic,
        every node near coord/.append style={
            font=\sffamily\fontsize{7}{8.4}\selectfont,
            inner sep=0pt,
            anchor=west,
            xshift=1pt,
        },
    ]
    \addplot[fill=kt15blue, area legend, bar shift=4pt] coordinates {
        (8.5,MADNet)  [8.5]
        (7.8,AoHNet)  [7.8]
        (6.8,ZOLE)    [6.8]
        (4.9,CST-Stereo) [4.9]
        (3.5,AdaStereo)  [3.5]
        (4.1,TS-PSMNet)  [4.1]
        (2.99,MAGIC-IGEV) [\textbf{2.99}]
    };
    \addplot[fill=kt12blue, area legend, bar shift=-4pt] coordinates {
        (9.3,MADNet)  [9.3]
        (8.6,AoHNet)  [8.6]
        (3.6,AdaStereo)  [3.6]
        (3.5,TS-PSMNet)  [3.5]
        (3.04,MAGIC-IGEV) [\textbf{3.04}]
    };
    \legend{KITTI 2015, KITTI 2012}
    \end{axis}
    \end{tikzpicture}

    \end{minipage}
\par\vspace{-7pt}
\noindent\begin{minipage}[t]{0.49\linewidth}
    \caption{\textbf{Qualitative comparison} on DrivingStereo~\citep{drivingstereo}. From top to bottom: sunny, cloudy, rainy,~and~foggy.}
    \label{fig: rainy_foggy}
\end{minipage}\hfill
\begin{minipage}[t]{0.49\linewidth}
    \caption{\textbf{Results} on KITTI training sets. 
    ZOLE~\citep{ZOLE} and CST-Stereo~\citep{cst-stereo} do not report KITTI 2012 results.}
    \label{fig: additional_domain_adaptation}
\end{minipage}
\end{figure}

After pre-training with geometric consistency, we evaluate unsupervised adaptation on KITTI 2015 and 2012 under both binocular and multi-baseline target-domain fine-tuning settings.

\textbf{Comparison on KITTI benchmarks.}
For a fair comparison with existing unsupervised methods, we adopt the standard binocular fine-tuning setting, in which the teacher and student observe identical stereo inputs.
As shown in~\cref{tab: kitti benchmark}, \method-IGEV$^{*}$ establishes a new state-of-the-art among unsupervised methods, surpassing the previous best method DualNet~\citep{dualnet}.
We attribute this to the strong generalization of our pre-trained model, which already outperforms several fine-tuned unsupervised methods without any target-domain fine-tuning.

\textbf{Extension to multi-baseline data.}
We simulate the multi-baseline setting by synthesizing novel views through warping and inpainting~\citep{zerostereo}.
As shown by \method-IGEV$^{\S}$ in~\cref{tab: kitti benchmark}, this setting achieves the lowest outlier rate, improving over the standard binocular setting.
These results show that \method benefits from additional target views during target-domain adaptation.

\begin{table}[t]
    
    \centering
    \caption{\textbf{Ablation studies} on component design. Row (a) is the SceneFlow-pretrained IGEVStereo baseline. ``$+$'' denotes continued training for 200K steps on the indicated data. We report the outlier rate in occluded regions on the KITTI 2015 training set. \dataset-CN: the clear-noon subset of \dataset. PTF: photometric threshold filtering. AMF: auto-masking filtering~\citep{monodepth2}. OAW: the proposed occlusion-aware weighting strategy for balancing supervision.}
    \label{tab:ablation}
    \par\vspace{2pt}
    \tablefont
    
    \begin{minipage}[b]{0.46\textwidth}
        \setlength{\tabcolsep}{2pt}
        \centering
        
        \begin{tabular}{rllc}
             \ctoprule{2-4}
             & Training Set & Supervision Signal & OCC \\
             \cmidrule{2-4}
             (a) & SceneFlow & ground-truth & 14.26 \\
             \cmidrule{2-4}
             (b) & + \dataset & ground-truth & \underline{10.75} \\
             (c) & + \dataset & photometric consistency & 97.44 \\
             (d) & + \dataset & binocular self-training & 15.55 \\
             \rowcolor{gray!15} (e) & + \dataset & \method & \textbf{10.37} \\
             (f) & + \dataset-CN & \method & 11.25 \\
             \cbottomrule{2-4}
        \end{tabular}
        \panelcaption{Training set vs. supervision signal}
        \label{tab: disentangle}

    \end{minipage}
    \hfill
    \begin{minipage}[b]{0.28\textwidth}
        \setlength{\tabcolsep}{2pt}
        \centering
        
        \begin{tabular}{rccccc}
             \ctoprule{2-6}
                  & $\mathcal{L}_{g}$ & $\mathcal{L}_{p}$ & $\mathcal{L}_{s}$ & EMA & OCC \\
             \cmidrule{2-6}
                (g) &           & \ding{51} & \ding{51}   &   & 97.44 \\
                (h) & \ding{51} & \ding{51} &             &  \ding{51} &  \underline{10.62} \\
                (i) & \ding{51} &            & \ding{51}  &  \ding{51} &    11.23 \\
                (j) & \ding{51} & \ding{51} & \ding{51}  & & 10.87 \\
                \rowcolor{gray!15} (k) & \ding{51} & \ding{51} & \ding{51} & \ding{51} &  \textbf{10.37} \\
             \cbottomrule{2-6}
        \end{tabular}
        \panelcaption{Loss terms and EMA teacher}
        \label{tab:ablation loss}

    \end{minipage}
    \hfill
    \begin{minipage}[b]{0.24\textwidth}
        \setlength{\tabcolsep}{1.5pt}
        \centering
        
        \begin{tabular}{rcccc}
        \ctoprule{2-5}
            & PTF & AMF & OAW & OCC \\
        \cmidrule{2-5}
            (l) &  &       &     & 13.67 \\
            (m) & \ding{51}   &   & & 13.49 \\
            (n) &  & \ding{51}  & & 13.27 \\
            (o) & \ding{51} & \ding{51} &  & \underline{12.39} \\
            \rowcolor{gray!15} (p) & \ding{51} & \ding{51} & \ding{51} &  \textbf{10.37} \\
        \cbottomrule{2-5}
        \end{tabular}
        \panelcaption{Filtering and weighting}
        \label{tab: ablation valid mask}

    \end{minipage}
\end{table}

\textbf{Comparison on KITTI training sets.}
Following prior works, we evaluate our models fine-tuned with multi-baseline data (\method-IGEV$^{\S}$) on the KITTI training sets (\cref{fig: additional_domain_adaptation}), outperforming TS-PSMNet~\citep{semi-stereo} and CST-Stereo~\citep{cst-stereo}.

\subsection{Ablation Study}
\label{sec: ablation}

\textbf{Training set vs. supervision signal.}
In~\cref{tab:ablation} (left), continued baseline training with ground-truth supervision on \dataset (a vs.\ b) markedly improves generalization, confirming the value of our dataset.
On the same \dataset training set, both photometric supervision (c) and standard binocular self-training (d) degrade relative to the baseline (a).
In contrast, \method (e) leverages visibility asymmetry to derive reliable occlusion supervision, achieving a lower occluded-region outlier rate than the ground-truth-supervised baseline (b).
Training on the clear-noon subset alone also improves occluded-region disparity estimation over the baseline (f vs.\ a), indicating that the gains cannot be attributed solely to the weather and illumination diversity of \dataset.

\textbf{Loss terms and EMA teacher.}
In the middle panel, removing the geometric consistency loss $\mathcal{L}_{g}$ (g) severely degrades occlusion performance because the remaining appearance-based terms provide no direct occlusion supervision.
With $\mathcal{L}_{g}$ retained, incorporating only the photometric term (h) or only the smoothness term (i) remains suboptimal, while jointly optimizing all three terms (k) achieves the best performance.
Comparing (j) and (k), the EMA teacher further improves performance by providing more stable and progressively refined pseudo-labels.

\textbf{Filtering and weighting.}
In the right panel, starting from the base variant (l), PTF (m) and AMF (n) improve supervision quality by filtering out uncertain pixels, while combining both (o) yields additional improvements.
Adding OAW (o vs.\ p) gives the largest gain in this ablation (12.39\% to 10.37\%) by strengthening sparse student-occluded yet teacher-visible supervision.

%% file: sec/5_conclusion.tex
\section{Conclusion}
We presented \method, a multi-baseline geometric consistency framework for learning disparity completion in occluded regions.
Visibility asymmetry enables the teacher to observe correspondences unavailable to the student, while baseline scaling aligns the teacher's disparity predictions to supervise the student's training.
Occlusion-aware weighting discards unreliable teacher guidance and strengthens sparse student-occluded yet teacher-visible supervision during learning.
We also introduced \dataset, a synthetic multi-baseline dataset spanning diverse scenes, weather, and illumination.
Pre-training on \dataset improves occlusion performance across backbones and generalization to real-world scenes and weather.
On KITTI, the pre-trained model already outperforms several fine-tuned unsupervised methods.
Further fine-tuning on binocular pairs achieves state-of-the-art unsupervised performance, with additional gains when more target views are available.
Future work will extend training to real multi-camera observations, scaling stereo learning without ground-truth disparity labels across diverse real-world scenes and challenging imaging conditions.

%% file: ai_use_statement.tex
\section*{AI Use Statement}

We used LLM-based assistants for literature retrieval and discovery, translation, and language polishing of author-written text. We also used these tools to write and edit plotting code and assist with LaTeX typesetting and figure and table layout. The methodology and experimental design were developed by the authors. LLM-based assistants were not used to generate manuscript drafts or synthetic datasets, implement the proposed method, or interpret experimental results. Assistance with mathematical proofs is not applicable to this work. The authors reviewed all AI-assisted work and take responsibility for the final content.

%% file: reproducibility_statement.tex
\section*{Reproducibility Statement}

The method and training objectives are described in Section~3, and the implementation details and evaluation protocols are provided in Section~4.1. Additional experimental and dataset details are included in the appendix. The source code will be publicly released upon acceptance.

%% file: sec/6_supp.tex
\nocite{aohnet,depthanything}
\noindent This document provides additional details and analyses supporting the main paper.
\cref{sec:impl} reports our computing infrastructure together with the implementation details omitted from the main paper.
\cref{sec:dataset} details our \dataset dataset, and \cref{sec:novel_view} introduces the construction of the multi-baseline fine-tuning data.
We then report quantitative results on the KITTI benchmarks in \cref{sec:kitti_supp}.
\cref{sec:weight_maps} quantifies the reliability of our cross-baseline pseudo-labels and the quality of the computed visibility masks.
\cref{sec:sensitivity} analyzes the sensitivity of our hyperparameters and the robustness of our framework to calibration error.
Finally, \cref{sec:future} outlines our plan to extend \method to multi-baseline observations captured with real camera arrays.

\section{Implementation Details}
\label{sec:impl}

\subsection{Computing Infrastructure}
All experiments are conducted on a single workstation equipped with four NVIDIA RTX 4090 GPUs (48\,GB of memory each), two Intel Xeon Gold 6148 CPUs at 2.40\,GHz, and 256\,GB of system memory, running Ubuntu 22.04.
Our implementation uses Python 3.10 and PyTorch 2.4.1.

\subsection{Auto-Masking Criterion}
\label{sec:automask}
Besides photometric thresholding, the visibility masks adopt the auto-masking criterion of~\citep{monodepth2}.
A pixel passes this criterion only when warping the target view lowers its photometric error relative to comparing the two images directly,
\begin{equation}
\mathcal{L}_p(I_{ref},\hat{I}_{ref}) < \mathcal{L}_p(I_{ref},I_{tgt}).
\label{eq: automask}
\end{equation}

The pixels this criterion removes differ between monocular depth estimation and stereo matching.
In the monocular setting the two images are consecutive video frames, and the criterion discards pixels whose appearance does not change with ego-motion, namely pixels in frames captured while the camera is static, and pixels on objects that move at the same velocity as the camera.
In stereo matching the two images are captured simultaneously by two cameras with a fixed baseline, so neither case arises.
A pixel instead fails~\cref{eq: automask} when its disparity is so small that warping leaves it essentially unchanged, or when its neighborhood is so uniform that the warped and unwarped errors are indistinguishable.
The former corresponds to points at very large depth, most notably sky.
The latter corresponds to textureless surfaces such as blank walls.
The supervision in both groups of pixels is unreliable, so they are excluded from supervision.

\subsection{Ablation Variants}
\label{sec:ablation_supp}
We detail the settings of variants (c) and (d) in Table 4 of the main paper, both following the same training protocol used for the full \method framework in row (e).

\textbf{Row (c).}
This variant continues training the SceneFlow-pretrained baseline~\citep{igev} on \dataset with the photometric loss and the edge-aware smoothness term~\citep{SsSnet} alone, dropping the geometric consistency loss and the teacher branch.
It is the same variant as row (g).
The photometric term is masked by $\mathcal{V}^s$ to discard erroneous supervision in occluded regions, leaving direct correspondence supervision only in non-occluded pixels.
We attribute this variant's degradation to catastrophic forgetting of the completion ability acquired during supervised pre-training.

\textbf{Row (d).}
This variant feeds the teacher and the student the same target view, so the baseline ratio becomes one and the geometric consistency loss reduces to conventional binocular self-training.
We also set $\omega=1$, since no student-occluded yet teacher-visible region arises.
The occlusion-aware weight map accordingly reduces to the teacher visibility mask $\mathcal{V}^t$, which still discards pseudo-labels where the teacher is occluded.
All other settings follow \method, including the EMA teacher, the student-only data augmentation, and the visibility filtering strategy.

\section{\dataset Dataset}
\label{sec:dataset}
\textbf{Data collection setup.}
We collect \dataset in the CARLA simulator~\citep{carla} using a Tesla Model 3 equipped with five synchronized RGB cameras.
Data collection spans seven towns (01, 02, 03, 05, 06, 07, and 10).
For each town, we first record driving trajectories at clear noon, then replay them under four additional conditions, namely clear sunset, clear night, cloudy noon, and rainy noon.
The ego-vehicle operates at speeds up to 30\,km/h, with frames captured at 0.5-second intervals.
All five cameras are rigidly mounted at a height of 2.26\,m, maintaining a uniform baseline of 0.5\,m between adjacent sensors to form a fixed multi-baseline array.

\textbf{Ground-truth depth acquisition.}
Although our training objective requires no ground-truth disparities, we record metric depth to support broader research uses.
A depth camera is co-located with the leftmost RGB camera, storing pixel-aligned depth measurements alongside the RGB images.
This inclusion supports fully supervised stereo training, monocular depth estimation, multi-task learning, and quantitative evaluations that require metric depth.
Following the standard CARLA configuration~\citep{carla}, depth maps are encoded as three-channel RGB images.
The metric depth $z$, measured in meters, is decoded from the three color channels as follows:
\begin{equation}
z = 1000 \times \frac{R + 256G + 256^2B}{256^3 - 1}.
\end{equation}

\textbf{Dataset statistics.}
\cref{tab: dataset stat.} summarizes the distribution of multi-baseline image sets collected across different towns and environmental conditions.
In total, \dataset comprises 20,417 synchronized multi-baseline sets.

\textbf{Input triplet sampling strategy.}
To maximize training diversity, we construct triplets from the $N$ synchronized views available at each timestamp using the following protocol:
(1) one image is uniformly selected as the \textit{reference view};
(2) two target views are independently sampled (with replacement) from the remaining $(N-1)$ images to serve as the student and teacher inputs.
This strategy generates $N (N-1)^2$ possible triplets per timestamp.
With our five-camera setup ($N=5$), this results in $5 \times 4^2 = 80$ unique triplets per frame.
Scaling this across the entire dataset yields a total of $20{,}417 \times 80 = 1{,}633{,}360$ training triplets.

\textbf{Handling targets on either side of the reference camera.}
Our sampling strategy allows the target camera to lie on either side of the reference camera, whereas the stereo network assumes a left reference image and a right target image.
When the target camera lies to the left, we horizontally flip both images and feed them to the network in reference--target order.
This reverses the horizontal correspondence direction, making the transformed pair compatible with the network's convention while preserving the identity of the reference view.
The predicted disparity map is then horizontally flipped back to align with the original reference image.
We represent disparity as a nonnegative magnitude and use positive baseline lengths; therefore, this final flip changes only the spatial indexing of the disparity map, without changing the sign or magnitude of any disparity value.

\begin{table}[t]
    
    \centering
    \caption{\textbf{Statistics of \dataset.} The dataset covers diverse weather and illumination conditions. Each set contains five color images and one ground-truth depth map.}
    \label{tab: dataset stat.}
    \par\vspace{7pt}
    \small
    \setlength{\tabcolsep}{2pt}
    \begin{tabular}{c|ccccc|c}
    \toprule
        Town & Clear Night & Clear Noon & Clear Sunset & Cloudy Noon & Rainy Noon & Total  \\
    \midrule
        01 & 630 & 630 & 629 & 632 & 631 & 3152  \\
        02 & 424 & 424 & 424 & 425 & 424 & 2121  \\
        03 & 647 & 649 & 648 & 649 & 648 & 3241  \\
        05 & 563 & 563 & 567 & 567 & 564 & 2824  \\
        06 & 820 & 821 & 820 & 821 & 821 & 4103  \\
        07 & 442 & 440 & 441 & 440 & 440 & 2203  \\
        10 & 552 & 557 & 554 & 554 & 556 & 2773  \\
    \midrule
        \textbf{Total} & \textbf{4078} & \textbf{4084} & \textbf{4083} & \textbf{4088} & \textbf{4084} & \textbf{20417} \\
    \bottomrule
    \end{tabular}
\end{table}

\begin{figure}[t]
    \centering
    \includegraphics[width=1\linewidth]{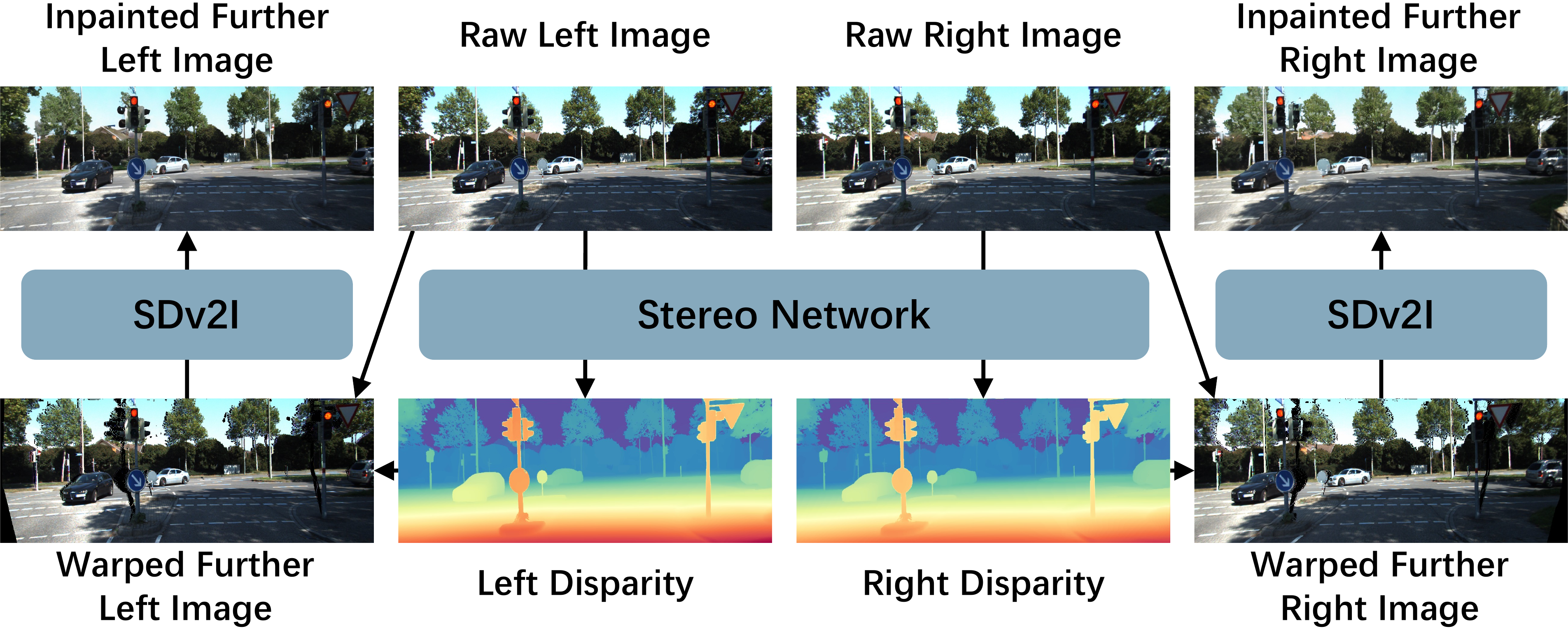}
    \par\vspace{-3pt}
    \caption{\textbf{Pipeline for constructing multi-baseline data from stereo image pairs.} SDv2I: Stable Diffusion v2 Inpainting model~\citep{zerostereo}.}
    \label{fig: inpainting}
\end{figure}

\section{Construction of Multi-Baseline Fine-Tuning Data}
\label{sec:novel_view}

Because KITTI 2012~\citep{KITTI2012} and KITTI 2015~\citep{KITTI2015} provide only binocular pairs, we synthesize additional views to evaluate multi-baseline fine-tuning in the target domain.
Following ZeroStereo~\citep{zerostereo}, we synthesize the additional views by warping and inpainting, but estimate disparity from each real stereo pair rather than from a monocular depth prior~\citep{depthanythingv2}.
The synthesized views are scale-consistent with the true camera baseline.

As illustrated in~\cref{fig: inpainting}, the view generation process consists of three steps:

\begin{enumerate}
    \item \textbf{Disparity estimation.}
    We use MonSter~\citep{monster} to estimate disparity maps for both images in each original left-right stereo pair.

    \item \textbf{Geometric warping.}
    Using the estimated disparity, we generate two additional outer views by warping the left image further to the left and the right image further to the right.
    The baseline between adjacent views matches the original stereo baseline since all warping operations are performed in the same original disparity coordinate space.

    \item \textbf{Generative inpainting.}
    Warping leaves disoccluded holes because some pixels in the warped view have no valid correspondence in the original view.
    Disparity errors near depth discontinuities can also cause boundary bleeding and distortion~\citep{adl}.
    We therefore dilate the occlusion mask to cover both kinds of artifacts and fill the resulting masked regions with an inpainting model~\citep{zerostereo}.
\end{enumerate}

The output contains four views per timestamp: the stereo pair and two generated outer views.
During fine-tuning, triplets sampled from these views provide supervision based on geometric consistency.

\textbf{Discussion.}
\method-IGEV$^{\S}$ relies on generated views as a proxy for real multi-baseline observations. Warping and inpainting artifacts in these generated views add noise to the supervision.
The reported gains concern this proxy setting. Real camera arrays introduce different acquisition errors, so the benefits of physically captured views cannot be inferred directly from these proxy results.

\section{Additional KITTI Benchmark Results}
\label{sec:kitti_supp}

\begin{table}[t]
    
    \centering
    \caption{
        \textbf{Additional results on the KITTI benchmarks}.
        This table extends Table 3 of the main paper with more evaluation metrics and baseline methods.
        The first block lists supervised methods, separated by a double rule from the unsupervised methods below.
        In the last block, the unmarked \method row denotes our pre-trained model without fine-tuning, $^{*}$ binocular fine-tuning, and $^{\S}$ multi-baseline fine-tuning, following the main paper.
        $^{\#}$ follows the settings of $^{\S}$ but trains from random initialization instead of the official SceneFlow-pretrained weights.
        \textbf{Bold} and \underline{underline} mark the best and second-best results among unsupervised methods only.
    }
    \label{tab: kitti benchmark supp}
    \par\vspace{7pt}
    \footnotesize
    \setlength{\tabcolsep}{1pt}
    \begin{tabular}{lcccccccccccc}
        \toprule
        \multirow{3}{*}{Method} & \multicolumn{6}{c}{KITTI 2015} & \multicolumn{6}{c}{KITTI 2012} \\
        \cmidrule(lr){2-7} \cmidrule(lr){8-13}
        & \multicolumn{3}{c}{NOC} & \multicolumn{3}{c}{ALL} & \multicolumn{3}{c}{NOC} & \multicolumn{3}{c}{ALL} \\
        \cmidrule(lr){2-4} \cmidrule(lr){5-7} \cmidrule(lr){8-10} \cmidrule(lr){11-13}
        & D1-BG & D1-FG & D1-ALL & D1-BG & D1-FG & {D1-ALL} & EPE & Out-2 & {Out-3} & EPE & Out-2 & Out-3 \\
        \midrule
        DispNet~(\citeyear{SceneFlow}) & 4.11 & 3.72 &  4.05 & 4.32 & 4.41 & 4.34 & 0.9 & 7.38 & 4.11 & 1.0 & 8.11 & 4.65\\
        SGM-Net~(\citeyear{SGM-Net}) & 2.23 & 7.44 & 3.09 & 2.66 &  8.64 & 3.66 & 0.7 & 3.60 & 2.29 & 0.9 & 5.15 & 3.50\\
        GCNet~(\citeyear{GCNet}) &  2.02 & 5.58  & 2.61 & 2.21 & 6.16 & 2.87 & 0.6 & 2.71 & 1.77  & 0.7 & 3.46 & 2.30 \\
        \midrule
        \midrule
        MADNet~(\citeyear{MADNet}) & 3.45 & 8.41 & 4.27 & 3.75 & 9.20 & 4.66 &  –– &  –– &  –– &  –– &  –– &  ––\\
        AdaStereo~(\citeyear{AdaStereo}) & 2.39 & 5.06 & 2.83 & 2.59 & 5.55 & 3.08 &  –– &  –– &  –– &  –– &  –– &  ––\\
        DualNet~(\citeyear{dualnet})  & 2.28 &  4.66 & 2.67 & 2.46 & 5.25 & 2.92 & \underline{0.6} & –– & 2.06  & \textbf{0.6} & –– & 2.59 \\
        MC-CNN-WS~(\citeyear{MC-CNN-WS})  & 3.06 & 9.42 & 4.11 & 3.78 & 10.93 & 4.97 & 0.8 & 4.76 & 3.02 & 1.0 & 6.57 & 4.45 \\
        SsSnet~(\citeyear{SsSnet})  & 2.46 & 6.13 & 3.06 & 2.70 & 6.92 & 3.40 & 0.7 & 3.34 & 2.30 & \underline{0.8} & 4.24 & 3.00 \\
        OASM~(\citeyear{OASM}) & 5.44 & 17.30 & 7.39  & 6.89 & 19.42 & 8.98 & 1.3 & 9.01 & 6.39  & 2.0 & 11.17 & 8.60\\
        Flow2Stereo~(\citeyear{flow2stereo}) & 4.77 & 14.03 & 6.29 & 5.01 & 14.62 & 6.61 & 1.0 & 6.56 & 4.58  & 1.1 & 7.32 & 5.11 \\
        Reversing-PSM~(\citeyear{reversing})  & 2.97 & 8.33 & 3.86 & 3.13 & 8.70 & 4.06 & –– & –– & –– & –– & –– & ––\\
        PASMnet~(\citeyear{PASMnet}) & 5.02 & 15.16 & 6.69  & 5.41 & 16.36 & 7.23 & 1.3 & –– & 7.14  & 1.5 & –– & 8.57\\
        EMR-MSF~(\citeyear{EMR-MSF}) & 8.30 & 14.16 & 9.27 & 8.61 & 15.15 & 9.70  & –– & –– & –– & –– & –– & ––\\
        UHP~(\citeyear{UHP}) & 4.65 & 12.37 & 5.93  & 5.00 & 13.70 & 6.45 & 1.2 & 9.08 & 6.05  & 1.3 & 10.37 & 7.09 \\
        \midrule
        \rowcolor{gray!15} \textbf{\method-IGEV} & 2.67 & 12.36 & 4.27 & 2.89 & 13.19 & 4.60 & 0.7 & 4.18 & 2.72 & \underline{0.8} & 5.05 & 3.27 \\
        \rowcolor{gray!15} \textbf{\method-IGEV}$^{*}$ & \underline{2.13} & 4.94 & \underline{2.59} & \underline{2.28} & 5.39 & \underline{2.79} & \underline{0.6} & \underline{3.23} & \underline{2.02} & \textbf{0.6} & \underline{3.79} & \underline{2.36} \\
        \rowcolor{gray!15} \textbf{\method-IGEV$^{\S}$} & \textbf{2.06} & \underline{4.43} & \textbf{2.45} & \textbf{2.21} & \underline{4.86} & \textbf{2.65} & \textbf{0.5} & \textbf{3.11} & \textbf{1.92} & \textbf{0.6} & \textbf{3.69} & \textbf{2.31} \\
        \rowcolor{gray!15} \textbf{\method-IGEV$^{\#}$} & 2.25 & \textbf{4.30} & \underline{2.59} & 2.44 & \textbf{4.75} & 2.82 & \underline{0.6} & 3.54 & 2.17 & \textbf{0.6} & 4.21 & 2.60 \\
        \bottomrule
        \end{tabular}
\end{table}

\begin{table}[t]
    
    \centering
    \caption{\textbf{Reliability of cross-baseline pseudo-labels.}
    Region (2) is teacher-occluded and region (3) is student-occluded yet teacher-visible, following the definitions in the method section of the main paper.
    We compare the rescaled teacher pseudo-label $r\cdot d^t$ against the student prediction $d^s$, both measured against the ground-truth disparity at the student baseline.
    Region (2) reports the teacher alone, since these pixels are discarded rather than compared.
    The upper block uses oracle visibility derived from ground-truth depth; the lower block uses photometric thresholding and auto-masking of fixed-model predictions, without training or weight updates.}
    \label{tab: vmask}
    \par\vspace{7pt}
    \small
    \begin{tabular}{lcccc}
    \toprule
        \multirow{2}{*}{Partition} & \multicolumn{2}{c}{Teacher pseudo-label} & \multicolumn{2}{c}{Student prediction} \\
        \cmidrule(lr){2-3} \cmidrule(lr){4-5}
         & EPE & Out-3 & EPE & Out-3 \\
    \midrule
        \multicolumn{5}{l}{\emph{Oracle visibility}} \\
        \quad Region (2) & 7.29 & 32.22 & –– & –– \\
        \quad Region (3) & \textbf{4.27} & \textbf{19.72} & 5.19 & 24.14 \\
    \midrule
        \multicolumn{5}{l}{\emph{Computed visibility}} \\
        \quad Region (2) & 7.66 & 28.23 & –– & –– \\
        \quad Region (3) & \textbf{2.27} & \textbf{12.91} & 3.52 & 17.75 \\
    \bottomrule
    \end{tabular}
\end{table}

As shown in~\cref{tab: kitti benchmark supp}, under binocular fine-tuning, \method-IGEV$^{*}$ surpasses all previous unsupervised methods on most reported metrics.
Binocular fine-tuning gives the teacher and the student the same target view, so we set $\omega=1$ as in row (d) and the occlusion-aware weight map reduces to the teacher visibility mask $\mathcal{V}^t$.
The official leaderboard does not report occluded-region metrics.
Compared with DualNet~\citep{dualnet}, however, our Out-3 on KITTI 2012~\citep{KITTI2012} decreases only marginally on NOC pixels, from 2.06\% to 2.02\%, but far more on ALL pixels, from 2.59\% to 2.36\%, which indicates that the gain concentrates on occluded pixels.
We attribute this to geometrically consistent pre-training across multiple baselines, which strengthens the ability to complete disparities within the occluded image regions.

\method-IGEV$^{\S}$ achieves the best results on most reported metrics, further supporting the use of multi-baseline observations for fine-tuning within the target domain.

\textbf{Random initialization.} To assess dependence on SceneFlow initialization, we also report \method-IGEV$^{\#}$ with random initialization.
As described in the main paper, photometric supervision anchors basic matching, teacher pseudo-labels cover regions where photometric consistency fails, and the EMA update propagates the student's improved disparity completion back to the teacher.
This complementary objective enables training from random initialization while remaining competitive with variants initialized from official SceneFlow-pretrained weights~\citep{SceneFlow}.

\section{Analysis of Occlusion-Aware Weighting}
\label{sec:weight_maps}

\begin{table}[t]
    
    \centering
    \caption{\textbf{Mask quality and region size} across the six camera pairs, all reported in percent.
    $\mathcal{V}^t$ precision is the fraction of pixels marked as teacher-visible by the computed mask that are genuinely teacher-visible, so its complement measures the fraction of admitted pixels that are actually teacher-occluded.
    Occlusion recall is the fraction of genuinely teacher-occluded pixels that $\mathcal{V}^t$ discards, so rows sharing the same $B^t$ report identical values.
    The last column additionally reports how much of the reference image falls in region (3), the pixels occluded to the student yet visible to the teacher.}
    \label{tab: mask quality}
    \par\vspace{7pt}
    \small
    \setlength{\tabcolsep}{5pt}
    \begin{tabular}{cccc}
    \toprule
        $B^t\!\rightarrow\!B^s$ (m) & $\mathcal{V}^t$ precision & Occ. recall & Region (3) \\
    \midrule
        $0.5\rightarrow1.0$ & 98.8 & 87.3 & 5.8 \\
        $0.5\rightarrow1.5$ & 98.8 & 87.3 & 9.8 \\
        $0.5\rightarrow2.0$ & 98.8 & 87.3 & 13.5 \\
        $1.0\rightarrow1.5$ & 98.5 & 91.7 & 5.5 \\
        $1.0\rightarrow2.0$ & 98.5 & 91.7 & 9.5 \\
        $1.5\rightarrow2.0$ & 98.4 & 93.6 & 5.5 \\
    \bottomrule
    \end{tabular}
\end{table}

Our framework exploits the premise that teacher-visible pixels provide more reliable predictions on average where the student is occluded.
We evaluate this premise using the pixel-aligned ground-truth depth maps recorded by the depth camera of our \dataset dataset.

\textbf{Setup.}
We evaluate 500 frames covering all five weather conditions at the full $540\times960$ resolution.
The reference view is fixed to the leftmost camera so that the co-located depth camera supplies ground-truth disparity, and we enumerate the six camera pairs satisfying $B^t<B^s$, the regime in which region (3) arises.
Oracle visibility is obtained by forward-warping the ground-truth disparity with a z-buffer test, which marks a pixel occluded when a nearer pixel maps to the same target pixel or when it falls outside the target view, both cases lacking a valid correspondence.
A single IGEVStereo model~\citep{igev} with official SceneFlow weights~\citep{SceneFlow} serves as both the teacher and the student, so the two branches differ only in their target view and the resulting baseline ratio, and the diagnostic involves no \method training.

\textbf{Teacher visibility yields more reliable supervision.}
As reported in~\cref{tab: vmask}, the teacher pseudo-label is more accurate than the student prediction on average in region (3), by 0.92\,px EPE under the oracle partition and by 1.25\,px under the computed partition.
The advantage holds even though rescaling by $r>1$ amplifies the teacher's error proportionally.
In region (2) the teacher pseudo-label is far less accurate than in region (3), with Out-3 above 28\% under both partitions, which supports discarding it.
Both partitions agree on the direction of every comparison, while photometric thresholding and auto-masking yield a computed partition with lower absolute errors.

\textbf{Mask quality and region size.}
As reported in~\cref{tab: mask quality}, the computed $\mathcal{V}^t$ has 98.4--98.8\% precision: only 1.2--1.6\% of the pixels admitted into $\mathcal{L}_g$ are actually teacher-occluded.
At the same time, $\mathcal{V}^t$ filters out 87.3--93.6\% of the genuinely teacher-occluded pixels.
Region (3) covers 5.5\% to 13.5\% of the reference image, a small fraction compared with the co-visible region.
The supervision unique to visibility asymmetry is therefore sparse, motivating the weight $\omega>1$.

\section{Sensitivity and Robustness}
\label{sec:sensitivity}

\begin{figure}[t]
    \centering
    \includegraphics[width=1\linewidth]{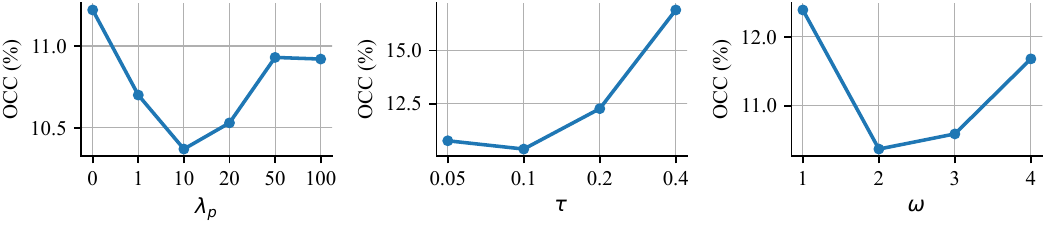}
    \par\vspace{-3pt}
    \caption{\textbf{Hyperparameter search.} We report Out-3 error in occluded regions on the KITTI 2015 training set~\citep{KITTI2015} while varying $\lambda_p$, $\tau$, and $\omega$.}
    \label{fig: ablation hyperparameter}
\end{figure}

\subsection{Hyperparameter Search}
\label{sec:hyperparameter_search}
We conduct a grid search over $\lambda_p$, $\tau$, and $\omega$, training on \dataset and evaluating on the KITTI 2015 training set~\citep{KITTI2015}.
As shown in~\cref{fig: ablation hyperparameter}, incorporating photometric supervision improves performance. However, excessively large $\lambda_p$ introduces noise due to imperfect occlusion masking.
Similarly, a loose threshold $\tau$ fails to adequately filter occluded pixels and leads to degraded performance.
For the occlusion-aware weighting factor $\omega$, a moderate value promotes disparity completion in student-occluded but teacher-visible regions, whereas an excessive value amplifies the noise in the teacher pseudo-labels.
We select $\lambda_p=10$, $\tau=0.1$, and $\omega=2$ using KITTI 2015 training-set annotations for hyperparameter selection only. We then evaluate cross-dataset generalization on KITTI 2012~\citep{KITTI2012}, Middlebury~\citep{Middlebury}, ETH3D~\citep{ETH3D}, and DrivingStereo~\citep{drivingstereo} with these hyperparameters fixed, without using annotations from these four datasets for training or hyperparameter selection.

\begin{figure}[t]
    \centering
    \includegraphics[width=1\linewidth]{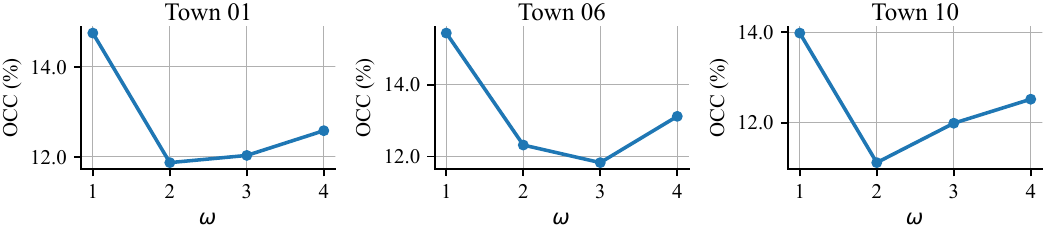}
    \par\vspace{-3pt}
    \caption{\textbf{Sensitivity of $\omega$ across scene layouts.} Each panel trains on a single town of \dataset and reports Out-3 error in occluded regions on the KITTI 2015 training set~\citep{KITTI2015} for different values of the occlusion-aware weighting parameter $\omega$.}
    \label{fig: omega town}
\end{figure}

\subsection{Sensitivity of \texorpdfstring{$\omega$}{omega} Across Scenes}
\label{sec:omega_scene}
The proportion of occluded pixels varies substantially across scenes, so a practical $\omega$ should generalize across them without retuning.
The search above trains on all towns of \dataset at once, so its $\omega$ reflects an average over layouts.
We therefore train on three towns with markedly different layouts, a small town (town 01), a rural road (town 06), and a high-rise city center (town 10), whose occlusion structures differ accordingly, and sweep $\omega$ alone.
As shown in~\cref{fig: omega town}, every value of $\omega$ between 2 and 4 outperforms $\omega=1$ in all three towns, and $\omega=2$ ranks best or second best in each.
These results support emphasizing the student-occluded yet teacher-visible region and using $\omega=2$ across different training layouts without retuning for each town.

\begin{figure}[t]
    \centering
    \includegraphics[width=1\linewidth]{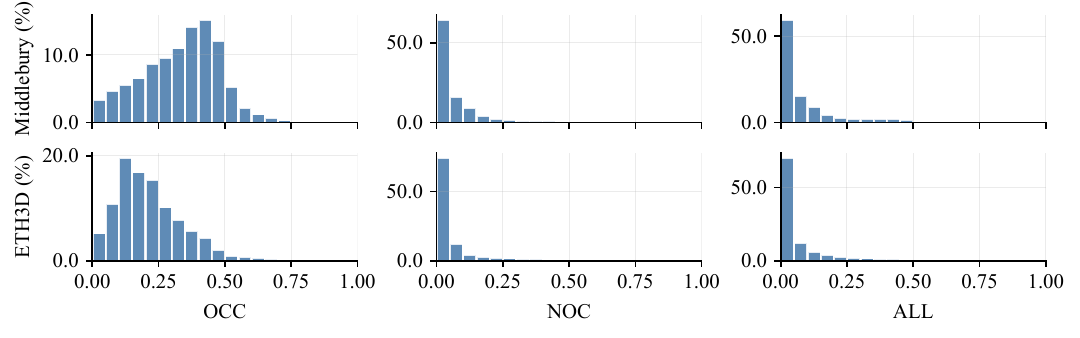}
    \par\vspace{-3pt}
    \caption{\textbf{Distribution of photometric error on real-world data.}
    We warp the right image to the left view with ground-truth disparity on Middlebury~\citep{Middlebury} and ETH3D~\citep{ETH3D} and histogram the resulting photometric error over occluded (OCC), non-occluded (NOC), and all (ALL) pixels.
    The horizontal axis is the photometric error and the vertical axis represents the percentage of pixels within each histogram bin.}
    \label{fig: photometric hist}
\end{figure}

\subsection{Transfer of the Photometric Threshold}
The threshold $\tau=0.1$ is used throughout pre-training on the synthetic \dataset, whose appearance differs substantially from real imagery.
Since the photometric error depends on appearance, we examine whether $\tau$ still separates occluded from non-occluded pixels on real data.
Middlebury~\citep{Middlebury} and ETH3D~\citep{ETH3D} provide ground-truth disparity and occlusion masks, which lets us warp the right image to the left view with ground-truth disparity and histogram the photometric error separately over the two populations.
As shown in~\cref{fig: photometric hist}, the two populations remain well separated on both datasets, with the non-occluded pixels concentrated near zero and the occluded pixels spread over much larger errors.
At $\tau=0.1$ the threshold retains 80.2\% and 86.1\% of the non-occluded pixels while filtering 92.1\% and 84.0\% of the occluded pixels on Middlebury and ETH3D, respectively, so the value used in synthetic pre-training transfers to both real indoor and outdoor imagery without retuning the photometric threshold.

\begin{table}[t]
    
    \centering
    \caption{\textbf{Robustness to baseline-ratio perturbation.} We perturb the baseline ratio during training as $r'=r(1+\epsilon)$ with $\epsilon\sim\mathcal{U}(-\alpha,\alpha)$ resampled for every training sample, and report Out-3 error on the KITTI 2015 training set in occluded (OCC), non-occluded (NOC), and all (ALL) regions.}
    \label{tab: calib}
    \par\vspace{7pt}
    \small
    \setlength{\tabcolsep}{6pt}
    \begin{tabular}{lccccc}
    \toprule
        $\alpha$ & 0 & 0.5\% & 1\% & 2\% & 5\% \\
    \midrule
        OCC & 10.37 & 10.27 & 10.46 & 10.53 & 11.14 \\
        NOC & 4.05 & 4.02 & 4.07 & 4.10 & 4.25 \\
        ALL & 4.21 & 4.18 & 4.23 & 4.27 & 4.42 \\
    \bottomrule
    \end{tabular}
\end{table}

\subsection{Robustness to Calibration Error}
\label{sec:calib}
Real multi-camera arrays carry extrinsic calibration error.
To verify how much baseline error our framework tolerates, we perturb the ratio $r=B^s/B^t$, the only quantity in the geometric consistency loss that depends on calibration.
During training, we sample $r' = r(1+\epsilon)$, where $\epsilon \sim \mathcal{U}(-\alpha, \alpha)$.
As reported in~\cref{tab: calib}, Out-3 in all three regions stays within 0.16 percentage points of the unperturbed setting for $\alpha$ up to 2\%.
At $\alpha=5\%$ the degradation remains graceful, with the occluded-region error rising from 10.37\% to 11.14\%.
The photometric term may contribute to this stability, since it anchors the student to appearance-based matching independently of the baseline ratio $r$.

\section{Future Work}
\label{sec:future}

All multi-baseline observations used in this work are either rendered in simulation or synthesized from binocular pairs.
The perturbation study of \cref{sec:calib} probes the sensitivity of \method to inaccurate baseline measurements, but it cannot replace physically captured data.
We therefore plan to build a real multi-baseline capture rig and collect a dataset spanning diverse scenes, illumination, and weather conditions.
Such data would let us study the acquisition artifacts that simulation does not reproduce, including inter-camera exposure and color mismatch as well as imperfect synchronization under ego-motion.
Our framework requires only prior calibration of the array baselines, and needs neither an expensive LiDAR sensor nor any other modality.
We aim to use low-cost camera arrays to close the synthetic-to-real gap and scale unsupervised stereo pre-training to real data.

%% file: magicup.bib
@String(ECCV= {Eur. Conf. Comput. Vis.})

@String(BMVC= {Brit. Mach. Vis. Conf.})

@String(ICLR = {Int. Conf. Learn. Represent.})

@String(AAAI = {AAAI})

@String(ECCV  = {ECCV})

@String(BMVC  =	{BMVC})

@String(ICLR  = {ICLR})

@inproceedings{depthanything,
  title={Depth anything: Unleashing the power of large-scale unlabeled data},
  author={Yang, Lihe and Kang, Bingyi and Huang, Zilong and Xu, Xiaogang and Feng, Jiashi and Zhao, Hengshuang},
  booktitle={Proceedings of the IEEE/CVF Conference on Computer Vision and Pattern Recognition},
  pages={10371--10381},
  year={2024}
}

@inproceedings{PSMNet,
  title={Pyramid stereo matching network},
  author={Chang, Jia-Ren and Chen, Yong-Sheng},
  booktitle={Proceedings of the IEEE conference on computer vision and pattern recognition},
  pages={5410--5418},
  year={2018}
}

@inproceedings{GCNet,
  title={End-to-end learning of geometry and context for deep stereo regression},
  author={Kendall, Alex and Martirosyan, Hayk and Dasgupta, Saumitro and Henry, Peter and Kennedy, Ryan and Bachrach, Abraham and Bry, Adam},
  booktitle={Proceedings of the IEEE international conference on computer vision},
  pages={66--75},
  year={2017}
}

@article{1,
  title={A taxonomy and evaluation of dense two-frame stereo correspondence algorithms},
  author={Scharstein, Daniel and Szeliski, Richard},
  journal={International journal of computer vision},
  volume={47},
  number={1},
  pages={7--42},
  year={2002},
  publisher={Springer}
}

@inproceedings{2,
  title={A closer look at memorization in deep networks},
  author={Arpit, Devansh and Jastrzebski, Stanis{\l}aw and Ballas, Nicolas and Krueger, David and Bengio, Emmanuel and Kanwal, Maxinder S and Maharaj, Tegan and Fischer, Asja and Courville, Aaron and Bengio, Yoshua and others},
  booktitle={International conference on machine learning},
  pages={233--242},
  year={2017},
  organization={PMLR}
}

@inproceedings{GwcNet,
  title={Group-wise correlation stereo network},
  author={Guo, Xiaoyang and Yang, Kai and Yang, Wukui and Wang, Xiaogang and Li, Hongsheng},
  booktitle={Proceedings of the IEEE/CVF Conference on Computer Vision and Pattern Recognition},
  pages={3273--3282},
  year={2019}
}

@inproceedings{CFNet,
  title={Cfnet: Cascade and fused cost volume for robust stereo matching},
  author={Shen, Zhelun and Dai, Yuchao and Rao, Zhibo},
  booktitle={Proceedings of the IEEE/CVF Conference on Computer Vision and Pattern Recognition},
  pages={13906--13915},
  year={2021}
}

@inproceedings{SceneFlow,
  title={A large dataset to train convolutional networks for disparity, optical flow, and scene flow estimation},
  author={Mayer, Nikolaus and Ilg, Eddy and Hausser, Philip and Fischer, Philipp and Cremers, Daniel and Dosovitskiy, Alexey and Brox, Thomas},
  booktitle={Proceedings of the IEEE conference on computer vision and pattern recognition},
  pages={4040--4048},
  year={2016}
}

@inproceedings{KITTI2015,
  title={Object scene flow for autonomous vehicles},
  author={Menze, Moritz and Geiger, Andreas},
  booktitle={Proceedings of the IEEE conference on computer vision and pattern recognition},
  pages={3061--3070},
  year={2015}
}

@inproceedings{KITTI2012,
  title={Are we ready for autonomous driving? the kitti vision benchmark suite},
  author={Geiger, Andreas and Lenz, Philip and Urtasun, Raquel},
  booktitle={2012 IEEE conference on computer vision and pattern recognition},
  pages={3354--3361},
  year={2012},
  organization={IEEE}
}

@inproceedings{MADNet,
  title={Real-time self-adaptive deep stereo},
  author={Tonioni, Alessio and Tosi, Fabio and Poggi, Matteo and Mattoccia, Stefano and Stefano, Luigi Di},
  booktitle={Proceedings of the IEEE/CVF Conference on Computer Vision and Pattern Recognition},
  pages={195--204},
  year={2019}
}

@article{3,
  title={Stereo matching by training a convolutional neural network to compare image patches.},
  author={Zbontar, Jure and LeCun, Yann and others},
  journal={J. Mach. Learn. Res.},
  volume={17},
  number={1},
  pages={2287--2318},
  year={2016}
}

@inproceedings{4,
  title={Efficient deep learning for stereo matching},
  author={Luo, Wenjie and Schwing, Alexander G and Urtasun, Raquel},
  booktitle={Proceedings of the IEEE conference on computer vision and pattern recognition},
  pages={5695--5703},
  year={2016}
}

@inproceedings{5,
  title={Patch Based Confidence Prediction for Dense Disparity Map.},
  author={Seki, Akihito and Pollefeys, Marc},
  booktitle={BMVC},
  volume={2},
  pages={4},
  year={2016}
}

@inproceedings{6,
  title={Sgm-nets: Semi-global matching with neural networks},
  author={Seki, Akihito and Pollefeys, Marc},
  booktitle={Proceedings of the IEEE conference on computer vision and pattern recognition},
  pages={231--240},
  year={2017}
}

@inproceedings{7,
  title={Improved stereo matching with constant highway networks and reflective confidence learning},
  author={Shaked, Amit and Wolf, Lior},
  booktitle={Proceedings of the IEEE conference on computer vision and pattern recognition},
  pages={4641--4650},
  year={2017}
}

@inproceedings{8,
  title={Detect, replace, refine: Deep structured prediction for pixel wise labeling},
  author={Gidaris, Spyros and Komodakis, Nikos},
  booktitle={Proceedings of the IEEE conference on computer vision and pattern recognition},
  pages={5248--5257},
  year={2017}
}

@inproceedings{CREStereo,
  title={Practical stereo matching via cascaded recurrent network with adaptive correlation},
  author={Li, Jiankun and Wang, Peisen and Xiong, Pengfei and Cai, Tao and Yan, Ziwei and Yang, Lei and Liu, Jiangyu and Fan, Haoqiang and Liu, Shuaicheng},
  booktitle={Proceedings of the IEEE/CVF Conference on Computer Vision and Pattern Recognition},
  pages={16263--16272},
  year={2022}
}

@inproceedings{RAFTStereo,
  title={Raft-stereo: Multilevel recurrent field transforms for stereo matching},
  author={Lipson, Lahav and Teed, Zachary and Deng, Jia},
  booktitle={2021 International Conference on 3D Vision (3DV)},
  pages={218--227},
  year={2021},
  organization={IEEE}
}

@inproceedings{AdaStereo,
  title={AdaStereo: a simple and efficient approach for adaptive stereo matching},
  author={Song, Xiao and Yang, Guorun and Zhu, Xinge and Zhou, Hui and Wang, Zhe and Shi, Jianping},
  booktitle={Proceedings of the IEEE/CVF Conference on Computer Vision and Pattern Recognition},
  pages={10328--10337},
  year={2021}
}

@inproceedings{Middlebury,
  title={High-resolution stereo datasets with subpixel-accurate ground truth},
  author={Scharstein, Daniel and Hirschm{\"u}ller, Heiko and Kitajima, York and Krathwohl, Greg and Ne{\v{s}}i{\'c}, Nera and Wang, Xi and Westling, Porter},
  booktitle={German conference on pattern recognition},
  pages={31--42},
  year={2014},
  organization={Springer}
}

@inproceedings{ETH3D,
  title={A multi-view stereo benchmark with high-resolution images and multi-camera videos},
  author={Schops, Thomas and Schonberger, Johannes L and Galliani, Silvano and Sattler, Torsten and Schindler, Konrad and Pollefeys, Marc and Geiger, Andreas},
  booktitle={Proceedings of the IEEE Conference on Computer Vision and Pattern Recognition},
  pages={3260--3269},
  year={2017}
}

@inproceedings{igev,
  title={Iterative Geometry Encoding Volume for Stereo Matching},
  author={Xu, Gangwei and Wang, Xianqi and Ding, Xiaohuan and Yang, Xin},
  booktitle={Proceedings of the IEEE/CVF Conference on Computer Vision and Pattern Recognition},
  pages={21919--21928},
  year={2023}
}

@inproceedings{dlnr,
  title={High-Frequency Stereo Matching Network},
  author={Zhao, Haoliang and Zhou, Huizhou and Zhang, Yongjun and Chen, Jie and Yang, Yitong and Zhao, Yong},
  booktitle={Proceedings of the IEEE/CVF Conference on Computer Vision and Pattern Recognition},
  pages={1327--1336},
  year={2023}
}

@inproceedings{adl,
  title={Adaptive Multi-Modal Cross-Entropy Loss for Stereo Matching},
  author={Xu, Peng and Xiang, Zhiyu and Qiao, Chengyu and Fu, Jingyun and Pu, Tianyu},
  booktitle={Proceedings of the IEEE/CVF Conference on Computer Vision and Pattern Recognition},
  pages={5135--5144},
  year={2024}
}

@article{depthanythingv2,
  title={Depth anything v2},
  author={Yang, Lihe and Kang, Bingyi and Huang, Zilong and Zhao, Zhen and Xu, Xiaogang and Feng, Jiashi and Zhao, Hengshuang},
  journal={Advances in Neural Information Processing Systems},
  volume={37},
  pages={21875--21911},
  year={2024}
}

@inproceedings{dualnet,
  title={DualNet: Robust Self-Supervised Stereo Matching with Pseudo-Label Supervision},
  author={Wang, Yun and Zheng, Jiahao and Zhang, Chenghao and Zhang, Zhanjie and Li, Kunhong and Zhang, Yongjian and Hu, Junjie},
  booktitle={Proceedings of the AAAI Conference on Artificial Intelligence},
  volume={39},
  number={8},
  pages={8178--8186},
  year={2025}
}

@article{BYOL,
  title={Bootstrap your own latent-a new approach to self-supervised learning},
  author={Grill, Jean-Bastien and Strub, Florian and Altch{\'e}, Florent and Tallec, Corentin and Richemond, Pierre and Buchatskaya, Elena and Doersch, Carl and Avila Pires, Bernardo and Guo, Zhaohan and Gheshlaghi Azar, Mohammad and others},
  journal={Advances in neural information processing systems},
  volume={33},
  pages={21271--21284},
  year={2020}
}

@inproceedings{monodepth2,
  title={Digging into self-supervised monocular depth estimation},
  author={Godard, Cl{\'e}ment and Mac Aodha, Oisin and Firman, Michael and Brostow, Gabriel J},
  booktitle={Proceedings of the IEEE/CVF international conference on computer vision},
  pages={3828--3838},
  year={2019}
}

@inproceedings{dino,
  title={Emerging properties in self-supervised vision transformers},
  author={Caron, Mathilde and Touvron, Hugo and Misra, Ishan and J{\'e}gou, Herv{\'e} and Mairal, Julien and Bojanowski, Piotr and Joulin, Armand},
  booktitle={Proceedings of the IEEE/CVF international conference on computer vision},
  pages={9650--9660},
  year={2021}
}

@inproceedings{drivingstereo,
  title={Drivingstereo: A large-scale dataset for stereo matching in autonomous driving scenarios},
  author={Yang, Guorun and Song, Xiao and Huang, Chaoqin and Deng, Zhidong and Shi, Jianping and Zhou, Bolei},
  booktitle={Proceedings of the IEEE/CVF conference on computer vision and pattern recognition},
  pages={899--908},
  year={2019}
}

@inproceedings{OASM,
  title={Occlusion aware stereo matching via cooperative unsupervised learning},
  author={Li, Ang and Yuan, Zejian},
  booktitle={Asian Conference on Computer Vision},
  pages={197--213},
  year={2018},
  organization={Springer}
}

@inproceedings{carla,
  title={CARLA: An open urban driving simulator},
  author={Dosovitskiy, Alexey and Ros, German and Codevilla, Felipe and Lopez, Antonio and Koltun, Vladlen},
  booktitle={Conference on robot learning},
  pages={1--16},
  year={2017},
  organization={PMLR}
}

@article{SSIM,
  title={Image quality assessment: from error visibility to structural similarity},
  author={Wang, Zhou and Bovik, Alan C and Sheikh, Hamid R and Simoncelli, Eero P},
  journal={IEEE transactions on image processing},
  volume={13},
  number={4},
  pages={600--612},
  year={2004},
  publisher={IEEE}
}

@inproceedings{defomstereo,
  title={DEFOM-Stereo: Depth Foundation Model Based Stereo Matching},
  author={Jiang, Hualie and Lou, Zhiqiang and Ding, Laiyan and Xu, Rui and Tan, Minglang and Jiang, Wenjie and Huang, Rui},
  booktitle={Proceedings of the IEEE/CVF Conference on Computer Vision and Pattern Recognition},
  pages={21857--21867},
  year={2025}
}

@inproceedings{foundationstereo,
  title={FoundationStereo: Zero-Shot Stereo Matching},
  author={Wen, Bowen and Trepte, Matthew and Aribido, Joseph and Kautz, Jan and Gallo, Orazio and Birchfield, Stan},
  booktitle={Proceedings of the IEEE/CVF Conference on Computer Vision and Pattern Recognition},
  pages={5249--5260},
  year={2025}
}

@inproceedings{wen2026fast,
  title={Fast-FoundationStereo: Real-time zero-shot stereo matching},
  author={Wen, Bowen and Dewan, Shaurya and Birchfield, Stan},
  booktitle={Proceedings of the IEEE/CVF Conference on Computer Vision and Pattern Recognition},
  pages={7513--7524},
  year={2026}
}

@inproceedings{monster,
  title={MonSter: Marry Monodepth to Stereo Unleashes Power},
  author={Cheng, Junda and Liu, Longliang and Xu, Gangwei and Wang, Xianqi and Zhang, Zhaoxing and Deng, Yong and Zang, Jinliang and Chen, Yurui and Cai, Zhipeng and Yang, Xin},
  booktitle={Proceedings of the IEEE/CVF Conference on Computer Vision and Pattern Recognition},
  pages={6273--6282},
  year={2025}
}

@inproceedings{wang2024selective,
  title={Selective-stereo: Adaptive frequency information selection for stereo matching},
  author={Wang, Xianqi and Xu, Gangwei and Jia, Hao and Yang, Xin},
  booktitle={Proceedings of the IEEE/CVF Conference on Computer Vision and Pattern Recognition},
  pages={19701--19710},
  year={2024}
}

@inproceedings{tosi2023nerf,
  title={Nerf-supervised deep stereo},
  author={Tosi, Fabio and Tonioni, Alessio and De Gregorio, Daniele and Poggi, Matteo},
  booktitle={Proceedings of the IEEE/CVF conference on computer vision and pattern recognition},
  pages={855--866},
  year={2023}
}

@inproceedings{multiscopic,
  title={Stereo Matching by Self-supervision of Multiscopic Vision},
  author={Yuan, Weihao and Zhang, Yazhan and Wu, Bingkun and Zhu, Siyu and Tan, Ping and Wang, Michael Yu and Chen, Qifeng},
  booktitle={2021 IEEE/RSJ International Conference on Intelligent Robots and Systems (IROS)},
  pages={5702--5709},
  year={2021}
}

@inproceedings{flow2stereo,
  title={Flow2stereo: Effective self-supervised learning of optical flow and stereo matching},
  author={Liu, Pengpeng and King, Irwin and Lyu, Michael R and Xu, Jia},
  booktitle={Proceedings of the IEEE/CVF conference on computer vision and pattern recognition},
  pages={6648--6657},
  year={2020}
}

@article{UHP,
  title={Unsupervised hierarchical iterative tile refinement network with 3D planar segmentation loss},
  author={Yang, Ruizhi and Li, Xingqiang and Cong, Rigang and Du, Jinsong},
  journal={IEEE Robotics and Automation Letters},
  volume={9},
  number={3},
  pages={2678--2685},
  year={2024},
  publisher={IEEE}
}

@inproceedings{MC-CNN-WS,
  title={Weakly supervised learning of deep metrics for stereo reconstruction},
  author={Tulyakov, Stepan and Ivanov, Anton and Fleuret, Francois},
  booktitle={Proceedings of the IEEE International Conference on Computer Vision},
  pages={1339--1348},
  year={2017}
}

@article{PASMnet,
  title={Parallax attention for unsupervised stereo correspondence learning},
  author={Wang, Longguang and Guo, Yulan and Wang, Yingqian and Liang, Zhengfa and Lin, Zaiping and Yang, Jungang and An, Wei},
  journal={IEEE transactions on pattern analysis and machine intelligence},
  volume={44},
  number={4},
  pages={2108--2125},
  year={2022},
  publisher={IEEE}
}

@article{SsSnet,
  title={Self-supervised learning for stereo matching with self-improving ability},
  author={Zhong, Yiran and Dai, Yuchao and Li, Hongdong},
  journal={arXiv preprint arXiv:1709.00930},
  year={2017}
}

@inproceedings{reversing,
  title={Reversing the cycle: self-supervised deep stereo through enhanced monocular distillation},
  author={Aleotti, Filippo and Tosi, Fabio and Zhang, Li and Poggi, Matteo and Mattoccia, Stefano},
  booktitle={European Conference on Computer Vision},
  pages={614--632},
  year={2020},
  organization={Springer}
}

@inproceedings{EMR-MSF,
  title={EMR-MSF: Self-Supervised Recurrent Monocular Scene Flow Exploiting Ego-Motion Rigidity},
  author={Jiang, Zijie and Okutomi, Masatoshi},
  booktitle={Proceedings of the IEEE/CVF International Conference on Computer Vision},
  pages={69--78},
  year={2023}
}

@inproceedings{adam,
  title={Adam: A Method for Stochastic Optimization},
  author={Kingma, Diederik P. and Ba, Jimmy},
  booktitle={International Conference on Learning Representations (ICLR)},
  year={2015},
  url={https://arxiv.org/abs/1412.6980}
}

@misc{onecyclelr,
      title={A disciplined approach to neural network hyper-parameters: Part 1 -- learning rate, batch size, momentum, and weight decay}, 
      author={Leslie N. Smith},
      year={2018},
      eprint={1803.09820},
      archivePrefix={arXiv},
      primaryClass={cs.LG},
      url={https://arxiv.org/abs/1803.09820}, 
}

@inproceedings{cst-stereo,
  title={Consistency-aware Self-Training for Iterative-based Stereo Matching},
  author={Zhou, Jingyi and Ye, Peng and Zhang, Haoyu and Yuan, Jiakang and Qiang, Rao and YangChenXu, Liu and Cailin, Wu and Xu, Feng and Chen, Tao},
  booktitle={Proceedings of the Computer Vision and Pattern Recognition Conference},
  pages={16641--16650},
  year={2025}
}

@inproceedings{ZOLE,
  title={Zoom and learn: Generalizing deep stereo matching to novel domains},
  author={Pang, Jiahao and Sun, Wenxiu and Yang, Chengxi and Ren, Jimmy and Xiao, Ruichao and Zeng, Jin and Lin, Liang},
  booktitle={Proceedings of the IEEE Conference on Computer Vision and Pattern Recognition},
  pages={2070--2079},
  year={2018}
}

@inproceedings{semi-stereo,
  title={Semi-stereo: A universal stereo matching framework for imperfect data via semi-supervised learning},
  author={Yue, Xin and Lu, Zongqing and Lin, Xiangru and Ren, Wenjia and Shao, Zhijing and Hu, Haonan and Zhang, Yu and Liao, Qingmin},
  booktitle={Proceedings of the IEEE/CVF Conference on Computer Vision and Pattern Recognition Workshops},
  pages={646--655},
  year={2024}
}

@inproceedings{zerostereo,
  title={Zerostereo: Zero-shot stereo matching from single images},
  author={Wang, Xianqi and Yang, Hao and Xu, Gangwei and Cheng, Junda and Lin, Min and Deng, Yong and Zang, Jinliang and Chen, Yurui and Yang, Xin},
  booktitle={Proceedings of the IEEE/CVF International Conference on Computer Vision},
  pages={28177--28187},
  year={2025}
}

@inproceedings{SGM-Net,
  title={Sgm-nets: Semi-global matching with neural networks},
  author={Seki, Akihito and Pollefeys, Marc},
  booktitle={Proceedings of the IEEE conference on computer vision and pattern recognition},
  pages={231--240},
  year={2017}
}

@inproceedings{aohnet,
  title={Faster self-adaptive deep stereo},
  author={Wang, Haiyang and Wang, Xinchao and Song, Jie and Lei, Jie and Song, Mingli},
  booktitle={Proceedings of the Asian Conference on Computer Vision},
  year={2020}
}

@InProceedings{zhang2024robusttransfer,
  author = {Zhang, Jiawei and Li, Jiahe and Huang, Lei and Yu, Xiaohan and Gu, Lin and Zheng, Jin and Bai, Xiao},
  title = {Robust Synthetic-to-Real Transfer for Stereo Matching},
  booktitle = {Proceedings of the IEEE/CVF Conference on Computer Vision and Pattern Recognition},
  year = {2024},
  pages = {20247--20257}
}

@InProceedings{ding2022kdmvs,
  author = {Ding, Yikang and Zhu, Qingtian and Liu, Xiangyue and Yuan, Wentao and Zhang, Haotian and Zhang, Chi},
  title = {{KD-MVS}: Knowledge Distillation Based Self-supervised Learning for Multi-view Stereo},
  booktitle = {Computer Vision -- ECCV 2022},
  year = {2022},
  pages = {630--646},
  publisher = {Springer}
}
